\documentclass[11pt]{article}
\PassOptionsToPackage{dvipsnames}{xcolor}  
\usepackage[preprint]{acl}
\usepackage{times}
\usepackage{latexsym}

\usepackage{amsmath}
\usepackage[nameinlink,capitalize,noabbrev]{cleveref} 
\usepackage{ulem} 
\usepackage{multirow}
\usepackage[T1]{fontenc}

\usepackage[utf8]{inputenc}

\usepackage{pifont}
\usepackage[most]{tcolorbox}
\usepackage{enumitem}
\usepackage{listings}

\definecolor{mygreen}{RGB}{0, 150, 0}
\definecolor{myred}{RGB}{200, 0, 0}
\definecolor{myblue}{RGB}{0, 0, 200}

\newcommand{\bluevar}[1]{\textcolor{blue}{<#1>}}
\newcommand{\pyvar}[1]{\texttt{\textcolor{red}{#1}}}
\definecolor{bg_instruction}{RGB}{240, 240, 245}
\definecolor{bg_context}{RGB}{255, 248, 240}
\definecolor{bg_response}{RGB}{240, 255, 240}

\newcommand{\hlgreen}[1]{\textcolor{mygreen}{\textbf{#1}}}
\newcommand{\hlred}[1]{\textcolor{myred}{\textbf{#1}}}

\lstdefinelanguage{json}{
    basicstyle=\ttfamily\scriptsize, 
    breaklines=true,
    keywords={true,false,null},
    keywordstyle=\color{blue}\bfseries,
    string=[b]",
    stringstyle=\color{red!70!black},
    comment=[l]{:},
    commentstyle=\color{black},
    columns=fullflexible,
    aboveskip=0pt, 
    belowskip=0pt  
}

\usepackage{booktabs} 
\usepackage{makecell} 
\usepackage{graphicx} 
\usepackage{tabularx}

\usepackage{microtype}

\usepackage{inconsolata}

\newcommand{\cmark}{\textcolor{ForestGreen}{\ding{51}}}
\newcommand{\xmark}{\textcolor{Red}{\ding{55}}}
\title{EnterpriseRAG: Benchmarking LLM Instruction Adherence and Robustness under Non-Ideal Enterprise Retrieval}

\author{
  Huiqi Miao, Xinbao Sun, Bo Wang, Fanyu Meng, Lijun Mei, Na Wu, Di Jin, Chao Deng, Junlan Feng \\
  Jiutian Research, China Mobile, Beijing, China \\
  \href{miaohuiqi,sunxinbao,wangbo@cmjt.chinamobile.com}{\{miaohuiqi,sunxinbao,wangbo\}@cmjt.chinamobile.com}
}

\begin{document}
\maketitle
\begin{abstract}
Enterprise RAG deployments face a critical reliability gap: while LLMs satisfy individual constraints at rates up to 84\%, only 27\% of responses meet all requirements simultaneously, revealing a 57-point orchestration gap. Existing benchmarks assume clean retrieval with simple queries, failing to capture production conditions where noisy documents and multi-dimensional constraints coexist.
We introduce EnterpriseRAG, a benchmark of 983 expert-validated samples across six domains that systematically simulates three failure modes absent from prior work: retrieval noise, knowledge gaps, and factual conflicts, coupled with complex instructions.
Evaluation of 13 state-of-the-art LLMs reveals a severe instruction adherence collapse, where high per-constraint satisfaction masks low holistic compliance. Critical findings expose deep barriers under knowledge gaps and factual conflicts, even with reasoning-enhanced inference, indicating production RAG requires explicit context-aware protocols and calibrated judgment. EnterpriseRAG provides a reproducible foundation for measuring and closing these gaps, directly informing deployment decisions for enterprise-scale RAG systems. 
We will release the benchmark and evaluation framework upon publication.

\end{abstract}


\section{Introduction}

Enterprise RAG systems face complex queries like \textit{"Summarize Q3 revenue by region in markdown tables. If data is incomplete, state 'Data Unavailable' rather than estimating. If audit and management reports conflict, cite both explicitly."} requiring simultaneous factual extraction, formatting compliance, and protocol adherence.

These challenges stem not from inadequate factual grounding, but from a fundamental evaluation gap. Current benchmarks assess RAG systems on clean retrieval scenarios with simple queries \citep{gao_enabling_2023,es_ragas_2025}, while production deployments face three compounding challenges absent from existing evaluations: \textbf{(1)} complex multi-constraint instructions integrating formatting rules with context-aware protocols for evidence adjudication; \textbf{(2)} high retrieval noise from latency-constrained systems that surface 10--20 documents with substantial irrelevant content; \textbf{(3)} frequent knowledge failures including coverage gaps and factual conflicts driven by temporal drift or source fallibility.

While recent work advances robustness testing \citep{zeng_rare_2025} and instruction following \citep{dong_toward_2024}, these efforts evaluate constraints in isolation with synthetic noise, missing the compounding complexity of real enterprise workflows where multiple dimensions interact.

We introduce \textbf{EnterpriseRAG}, a benchmark grounded in real-world enterprise deployments, comprising 983 expert-validated samples across six vertical domains. Unlike prior benchmarks overlaying synthetic instructions onto standard datasets, EnterpriseRAG reflects authentic multi-domain scenarios derived from real operational queries. We preserve original user intents while systematically scaling up constraint complexity through an expert-informed synthesis protocol, and construct three orthogonal non-ideal retrieval modes (irrelevant noise, knowledge gaps, and factual conflicts) validated through LLM-assisted generation and human verification.
\begin{table*}[t]
\centering
\caption{Comparison with prior RAG benchmarks across Source, Complexity, and Robustness.}
\label{tab:rag-benchmarks-comparison}
\resizebox{0.85\textwidth}{!}{%
    \begin{tabular}{lcccccc}
    \toprule
    {\color[HTML]{1F1F1F} \textbf{Benchmark}} &
      \multicolumn{3}{c}{{\color[HTML]{1F1F1F} \textbf{Dataset Source}}} &
      \multicolumn{1}{c}{{\color[HTML]{1F1F1F} \textbf{\begin{tabular}[c]{@{}c@{}}Task\\ Complexity\end{tabular}}}} &
      \multicolumn{2}{c}{{\color[HTML]{1F1F1F} \textbf{Robustness}}} \\ \hline
     &
      {\color[HTML]{1F1F1F} \textit{\begin{tabular}[c]{@{}l@{}}Human-\\ Curated\end{tabular}}} &
      {\color[HTML]{1F1F1F} \textit{\begin{tabular}[c]{@{}l@{}}Vertical\\ Domain\end{tabular}}} &
      {\color[HTML]{1F1F1F} \textit{\begin{tabular}[c]{@{}l@{}}Natural User\\ Queries\end{tabular}}} &
      {\color[HTML]{1F1F1F} \textit{\begin{tabular}[c]{@{}l@{}}Complex \\ Constraints\end{tabular}}} &
      {\color[HTML]{1F1F1F} \textit{\begin{tabular}[c]{@{}l@{}}Negative \\ Rejection\end{tabular}}} &
      {\color[HTML]{1F1F1F} \textit{Conflict}} \\ \hline
    {\color[HTML]{1F1F1F} CRUD-RAG\citep{lyu_crud-rag_2024}} &
      \xmark & \cmark & \xmark & \xmark & \xmark & \xmark \\
    {\color[HTML]{1F1F1F} CRAG\citep{yang_crag_2024}} &
      \cmark & \cmark & \xmark & \xmark & \cmark & \xmark \\
    {\color[HTML]{1F1F1F} RAGBench\citep{friel_ragbench_2025}} &
      \xmark & \cmark & \cmark & \xmark & \xmark & \xmark \\
    {\color[HTML]{1F1F1F} RAGEval\citep{zhu_rageval_2025}} &
      \xmark & \cmark & \xmark & \xmark & \xmark & \xmark \\
    {\color[HTML]{1F1F1F} FollowRAG\citep{dong_toward_2024}} &
      \cmark & \xmark & \xmark & \cmark & \xmark & \xmark \\
    {\color[HTML]{1F1F1F} EKRAG\citep{yu_ekrag_2025}} &
      \cmark & \cmark & \cmark & \xmark & \xmark & \xmark \\
    {\color[HTML]{1F1F1F} RARE\citep{zeng_rare_2025}} &
      \xmark & \cmark & \xmark & \xmark & \cmark & \cmark \\
    {\color[HTML]{1F1F1F} GaRaGe\citep{sorodoc_garage_2025}} &
      \cmark & \cmark & \xmark & \xmark & \cmark & \xmark \\ \hline
    {\color[HTML]{1F1F1F} \textbf{EnterpriseRAG (Ours)}} &
      \cmark & \cmark & \cmark & \cmark & \cmark & \cmark \\ 
    \bottomrule
    \end{tabular}%
}
\end{table*}
Our evaluation framework extends traditional RAG metrics with \textbf{Strict IAS} (holistic compliance) versus \textbf{Loose IAS} (per-constraint satisfaction) to expose compositional adherence failures, plus robustness indicators for safety-critical scenarios. Testing 13 state-of-the-art LLMs reveals that while models handle structural formatting adequately, they systematically fail on behavioral protocols, particularly judgment under uncertainty, where even reasoning models achieve insufficient reliability for production deployment.

\paragraph{Contributions.}
Our contributions are threefold:
\begin{itemize}[leftmargin=*,nosep]
    \item \textbf{Enterprise-grade RAG benchmark}: We introduce \textbf{EnterpriseRAG}, 983 expert-validated instances across six domains, pairing complex multi-constraint instructions with three controlled non-ideal retrieval settings, plus a reproducible construction pipeline.
    \item \textbf{Comprehensive evaluation framework}:We develop specialized metrics for non-ideal contexts: \textbf{Loose/Strict IAS} for instruction adherence, and \textbf{rejection/conflict accuracy} to measure safety-critical judgment in scenarios with missing or conflicting information.
    \item \textbf{Experimental Insights}: Across 13 LLMs, we find an orchestration gap of up to 57pp (83.8\% Loose vs.\ 26.8\% Strict), with best rejection accuracy only 42.7\% and conflict recognition <45\%, indicating behavioral judgment under uncertainty as the main bottleneck for enterprise RAG.
\end{itemize}


\section{Related Work}
RAG evaluation has evolved from factual correctness toward robustness and instruction compliance \citep{zhao2024retrievalaugmentedgenerationaigeneratedcontent,gupta2024comprehensivesurveyretrievalaugmentedgeneration}. Table~\ref{tab:rag-benchmarks-comparison} positions EnterpriseRAG among representative benchmarks. A recurring pattern across prior work is that retrieval quality and instruction adherence are studied separately—leaving open whether models can satisfy both under realistic enterprise conditions.


\noindent\textbf{RAG Evaluation Paradigms.} Early benchmarks focused on factual accuracy and grounding (ALCE \citep{gao_enabling_2023}, RAGAS \citep{es_ragas_2025}) or multi-hop reasoning \citep{tang_multihop-rag_2024}. While domain-specific benchmarks exist (\citep{jin_pubmedqa_2019};\citep{pipitone_legalbench-rag_2024};\citep{chen_disc-finllm_nodate}), they often rely on curated sources like Wikipedia \citep{yang2018hotpotqadatasetdiverseexplainable}. Recent frameworks like RAGBench \citep{friel_ragbench_2025}, RAGEval \citep{zhu_rageval_2025} and EKRAG \citep{yu_ekrag_2025} offer multi-dimensional metrics and human-curated enterprise samples but lack systematic assessment of complex instruction adherence.

\noindent\textbf{Non-Ideal Retrieval Contexts.} Benchmarks like CRAG \citep{yang_crag_2024}, CRUD-RAG \citep{lyu_crud-rag_2024} and GaRaGe \citep{sorodoc_garage_2025} address dynamic KBs and calibration. Others such as RGB \citep{chen2023benchmarkinglargelanguagemodels}, RARE \citep{zeng_rare_2025}, and Magic Mushroom \citep{zhang2025magicmushroomcustomizablebenchmark} introduce noise and unanswerability. While conflict \citep{lee2025magicmultihopgraphbasedbenchmark,choi-etal-2025-conflict} and gap management \citep{guo2025llmcentricragmultigranularindexing} exist, they are typically evaluated in isolation, decoupled from uncertainty protocols.

\begin{figure*}[t]
  \includegraphics[width=\textwidth]{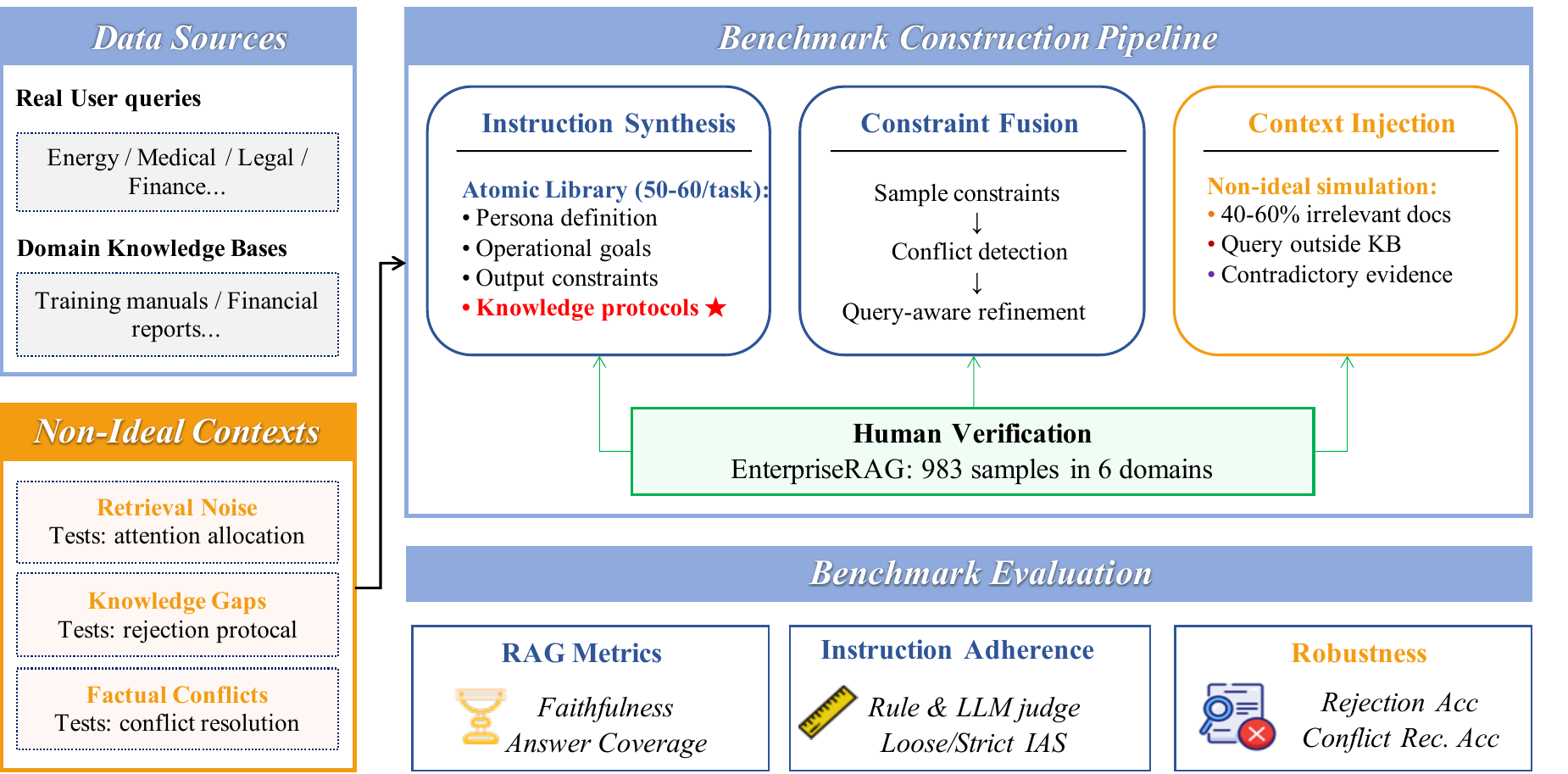}
  \caption{EnterpriseRAG Overview. The top section presents the pipeline for complex instruction schema design, while the bottom section shows non-ideal context simulation and evaluation metrics, respectively. All are automatically generated by LLM and quality verified by humans.}
  \label{fig:overview}
\end{figure*}

\noindent\textbf{Instruction Following in RAG.} General benchmarks \citep{zhou_instruction-following_2023,jiang-etal-2024-followbench,wen_benchmarking_2024,zou_eifbench_2025,qin_sysbench_2024} highlight the need for strict verification of formatting and negative constraints. In RAG contexts, MT-RAG \citep{katsis_mtrag_2025} and CRP-RAG \citep{xu2025crp} overlook strict protocol adherence, while FollowRAG \citep{dong_toward_2024} remains limited by synthetic injection and clean contexts. EnterpriseRAG targets the intersection: complex multi-constraint instructions grounded in operational workflows, evaluated under systematic retrieval noise, knowledge gaps, and factual conflicts.

\section{EnterpriseRAG Benchmark Construction}
\label{sec:enterpriseRAG_construction}

We construct EnterpriseRAG from production RAG logs, yielding 983 expert-validated instances spanning six domains (\textit{Energy, Medical, Legal, Financial, Party Building and Web Search}). 
All data are desensitized to remove PII; raw logs cannot be released, but we will release the desensitized benchmark and a reproducible generation pipeline (details in Appendix~\ref{subsec:appendix A.1}).

\paragraph{Instance format.}
Each instance is a triple $\langle q, \mathcal{I}, \mathcal{D}\rangle$: user query $q$, a fused multi-constraint instruction set $\mathcal{I}$, and a retrieved document bundle $\mathcal{D}$.
Starting from 491 authentic queries, we construct 983 instances by pairing queries with controlled non-ideal retrieval scenarios (overview in Figure~\ref{fig:overview}; subset composition in Table~\ref{tab:Nonideal_distribution}). Figure \ref{fig:legal_full_case} illustrates a Legal domain case study.

\paragraph{Construction pipeline.}
Our pipeline operationalizes two principles: \textbf{realistic constraint complexity} (from enterprise prompts) and \textbf{controlled non-ideal retrieval} (noise/gap/conflict).
We: (1) collect and filter queries; (2) synthesize $\mathcal{I}$ by fusing atomic constraints and removing internal contradictions; (3) retrieve documents via hybrid retrieval (BM25+dense) and assign non-ideal modes; (4) conduct expert verification for instruction--query consistency and context validity (Appendix~\ref{subsec:appendix A.2}). Prompt templates and quality control recipes are documented in Appendix~\ref{subsec:appendix D.1}.
\begin{table*}[t]
\centering
\caption{Performance of various LLMs on noisy subset. The average scores across the dataset are reported as percentages. The best and second-best scores are marked in \textbf{bold} and \underline{underlined}, respectively.}
\label{tab:standard-subset}
\resizebox{\textwidth}{!}{%
    \begin{tabular}{lcccccccc}
    \toprule
    \multirow{2}{*}{\textbf{Model}} & \multirow{2}{*}{\begin{tabular}[c]{@{}c@{}}Inference\\ Paradigm\end{tabular}} & \multicolumn{2}{c}{RAG Quality} & \multicolumn{2}{c}{IAS} & \multicolumn{3}{c}{Loose IAS} \\ 
    \cmidrule(lr){3-4} \cmidrule(lr){5-6} \cmidrule(lr){7-9}
    & & Faithfulness & \begin{tabular}[c]{@{}c@{}}Answer\\ Coverage\end{tabular} & Loose & Strict & \begin{tabular}[c]{@{}c@{}}Persona\\ Definition\end{tabular} & \begin{tabular}[c]{@{}c@{}}Output\\ Constraints\end{tabular} & \begin{tabular}[c]{@{}c@{}}Knowledge\\ Interaction\\ Protocol\end{tabular} \\ 
    \midrule
    \multicolumn{9}{c}{\textit{open-source models}} \\
    \midrule
    Qwen3-8b & Reasoning & 64.8 & 56.4 & 75.5 & 12.3 & 70.1 & 82.0 & 70.3 \\
    Qwen3-14b & Reasoning & 66.8 & 59.4 & 77.0 & 14.3 & 74.2 & 81.5 & 72.4 \\
    Qwen3-32b & Reasoning & 64.9 & 59.5 & 77.2 & 15.9 & 75.4 & 80.1 & 75.1 \\
    Qwen3-235B-A22B-Thinking-2507 & Reasoning & 67.1 & 64.1 & \textbf{83.8} & \textbf{26.8} & 82.2 & \textbf{86.2} & \underline{82.5} \\
    Qwen3-30B-A3B-Instruct-2507 & Standard & 63.9 & 65.6 & 76.4 & 13.2 & 79.0 & 81.5 & 68.9 \\
    Qwen3-235B-A22B-Instruct-2507 & Standard & 67.4 & \underline{67.1} & 80.6 & 20.8 & 82.3 & 83.6 & 76.3 \\
    DeepSeek-R1-0528 & Reasoning & 68.9 & 66.4 & 83.1 & 21.9 & 81.3 & 84.3 & \textbf{82.9} \\
    DeepSeek-V3.1 & Standard & 69.9 & 60.4 & 82.2 & 22.1 & 79.9 & \underline{85.9} & 78.8 \\
    GLM-4.5 & Reasoning & \underline{76.6} & 64.3 & 81.6 & 21.5 & 75.4 & 85.7 & 78.7 \\ 
    \midrule
    \multicolumn{9}{c}{\textit{closed-source models}} \\
    \midrule
    Gemini-2.5-Pro & Reasoning & 73.5 & 61.6 & \underline{83.7} & \underline{26.5} & \textbf{86.6} & 85.6 & 80.3 \\
    GPT-4.1 & Standard & 69.8 & 65.8 & 80.0 & 19.5 & 79.4 & 82.1 & 77.8 \\
    Claude-Opus-4.5 & Reasoning & 76.4 & \textbf{68.5} & 83.3 & 25.3 & \underline{84.7} & 84.1 & 82.4 \\
    Claude-Sonnet-4 & Standard & \textbf{76.8} & 66.7 & 79.8 & 19.5 & 77.1 & 81.5 & 79.1 \\ 
    \bottomrule
    \end{tabular}%
}
\end{table*}

\subsection{Complex Instruction Schema}
\label{subsec:instruction_schema_main}
We organize constraints into three orthogonal dimensions: \textbf{Persona Definition}, \textbf{Output Constraints}, and \textbf{Knowledge Interaction Protocols}.
This schema captures enterprise-critical behavioral requirements (e.g., citation, gap identification, conflict handling) in addition to structural formatting rules.
Definitions and distributions are provided in Appendix~\ref{subsec:appendix A.3}--\ref{subsec:appendix A.4} (Table~\ref{tab:standard-subset-subdim}, Table~\ref{tab:data_distribution}, Figure~\ref{fig:subdimension_constraint_distribution}).

\begin{figure}[t!]
    \centering
  \includegraphics[width=\columnwidth]{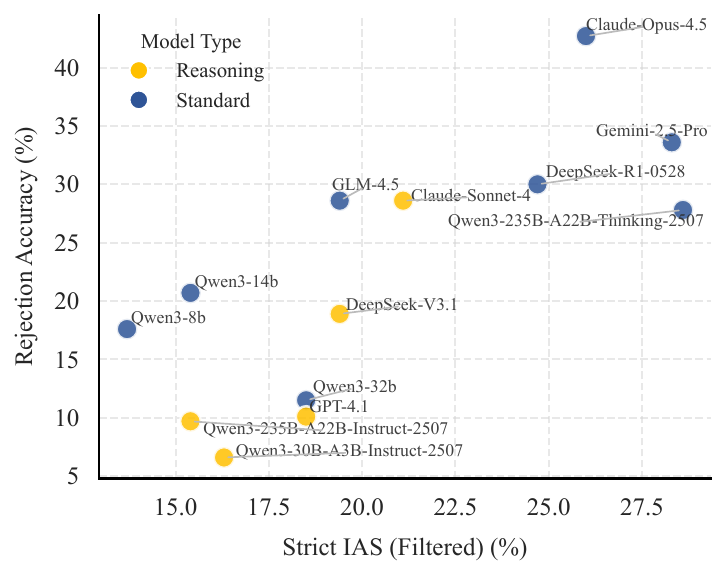}
  \caption{Instruction Adherence and Robustness comparison on the knowledge gap subset. Reasoning-enhanced models generally demonstrate superior capability in both protocol adherence and refusal of unanswerable queries.}
  \label{fig:reject_acc}
\end{figure}

\begin{figure}[t]
\centering
\includegraphics[width=0.48\textwidth]{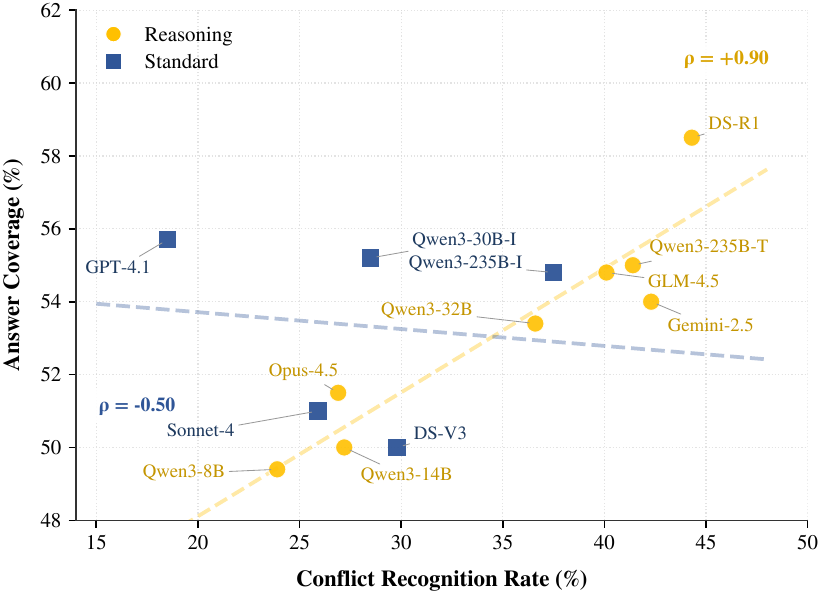}
\caption{Correlation between Conflict Recognition Rate and Answer Coverage 
across 13 models on the factual conflict subset (309 cases). Each point 
represents one model. Reasoning-enhanced models exhibit a 
strong positive correlation ($\rho$ = +0.90), while standard models 
show no significant relationship ($\rho$ = -0.50).}
\label{fig:conflict_coverage_corr}
\end{figure}

\subsection{Non-Ideal Retrieval Scenarios}
\label{subsec:nonideal_main}

We construct three non-ideal retrieval scenarios that stress the generator under realistic enterprise failure conditions:
\textbf{Noisy Retrieval} (topically similar but contextually irrelevant documents), \textbf{Knowledge Gaps} (topically related but insufficient evidence in retrieved contexts), and \textbf{Factual Conflicts} (contradictory statements in retrieved passages).
Because gaps and conflicts are sparse in natural logs, we augment them with controlled procedures while preserving domain coherence; we further validate that synthetic conflicts match natural difficulty on core robustness signals (Appendix~\ref{subsec:appendix A.5} and Appendix~\ref{subsec:appendix B.2}).

\subsection{Evaluation Metrics}
\label{subsec:Evaluation_Metrics}


Given the open-ended nature of enterprise queries, we report RAG quality and instruction adherence signals without requiring gold reference answers.
Faithfulness and Answer Coverage follow a RAGAS-style claim-based evaluation \citep{es_ragas_2025}.

\noindent\textbf{Faithfulness ($\mathcal{F}$).} We calculate faithfulness as $\mathcal{F} = |C_{sup}| / |C_{total}|$,
where $C_{total}$ denotes all claims extracted from the response, and $C_{sup}$ denotes those supported by the retrieved context.

\paragraph{Answer Coverage (C).}
\begin{equation}
  \label{eq:coverage_app}
  \mathcal{C} = \alpha \frac{|C_{ans} \cap R|}{|C_{ans}|} + (1-\alpha) \frac{|S_{ans} \cap R|}{|S_{ans}|}
\end{equation}
where $C_{ans}$ and $S_{ans}$ denote core and supplementary claims from contexts, respectively, $R$ represents the response content, and $\alpha=0.7$ weights core claims higher.

\paragraph{Instruction Adherence Score (IAS).}
We report:
\textbf{Loose IAS} as the proportion of satisfied constraints, and
\textbf{Strict IAS} as a binary score indicating whether \emph{all} constraints are satisfied.

\paragraph{Robustness metrics.}
For non-ideal subsets, we compute:
\textit{Rejection Accuracy} on knowledge gaps, and
\textit{Conflict Recognition Accuracy} on factual conflicts.


\section{Experiments}

\subsection{Experimental Setup}

\textbf{Models.} We evaluate 13 LLMs spanning open/closed-source and standard/reasoning-enhanced variants \citep{yang_qwen3_2025,deepseek-ai_deepseek-r1_2025,gpt_4_1,team_glm-45_2025,comanici_gemini_2025,Claude_Opus_4_5_model_card}, full list in Table~\ref{tab:standard-subset}. For consistency, we adopt simplified names after the first mention: ``Thinking'' models are denoted as \textit{-Thinking} (abbrev.\ \textit{-T}) and standard instruction-tuned counterparts as \textit{-Instruct} (abbrev.\ \textit{-I}).
We omit version suffixes (e.g., \textit{-2507}) unless needed for disambiguation.

\paragraph{Evaluation protocol.}
Results are reported on three non-ideal subsets: Noisy Retrieval ($n{=}447$), Knowledge Gaps ($n{=}227$), and Factual Conflicts ($n{=}309$).
IAS evaluation uses rule-based checks for structural constraints~\citep{zhou_instruction-following_2023} and LLM-as-a-judge (Kimi-k2-thinking~\citep{team_kimi_2025}) for behavioral protocols. Not all instances include explicit Knowledge Interaction Protocol constraints; we therefore compare naturally protocol-present vs.\ protocol-absent cases for robustness analyses.
Evaluation prompt templates are in Appendix~\ref{sec:appendix D}.

\paragraph{Evaluator reliability.}
Cross-judge comparison across three LLM evaluators shows stable scores and consistent model rankings (\Cref{subsec:appendix B.1}).
On 150 human-annotated samples, experts achieve strong agreement ($\kappa{=}0.85$), and the LLM evaluator (Kimi-k2-thinking) aligns well with human-annotated gold labels ($\kappa{=}0.77$, 88\% agreement), with especially high alignment on conflict recognition ($\kappa{=}0.93$; \Cref{subsec:appendix B.3}).

\begin{figure}[t]
    \centering
    \includegraphics[width=\columnwidth]{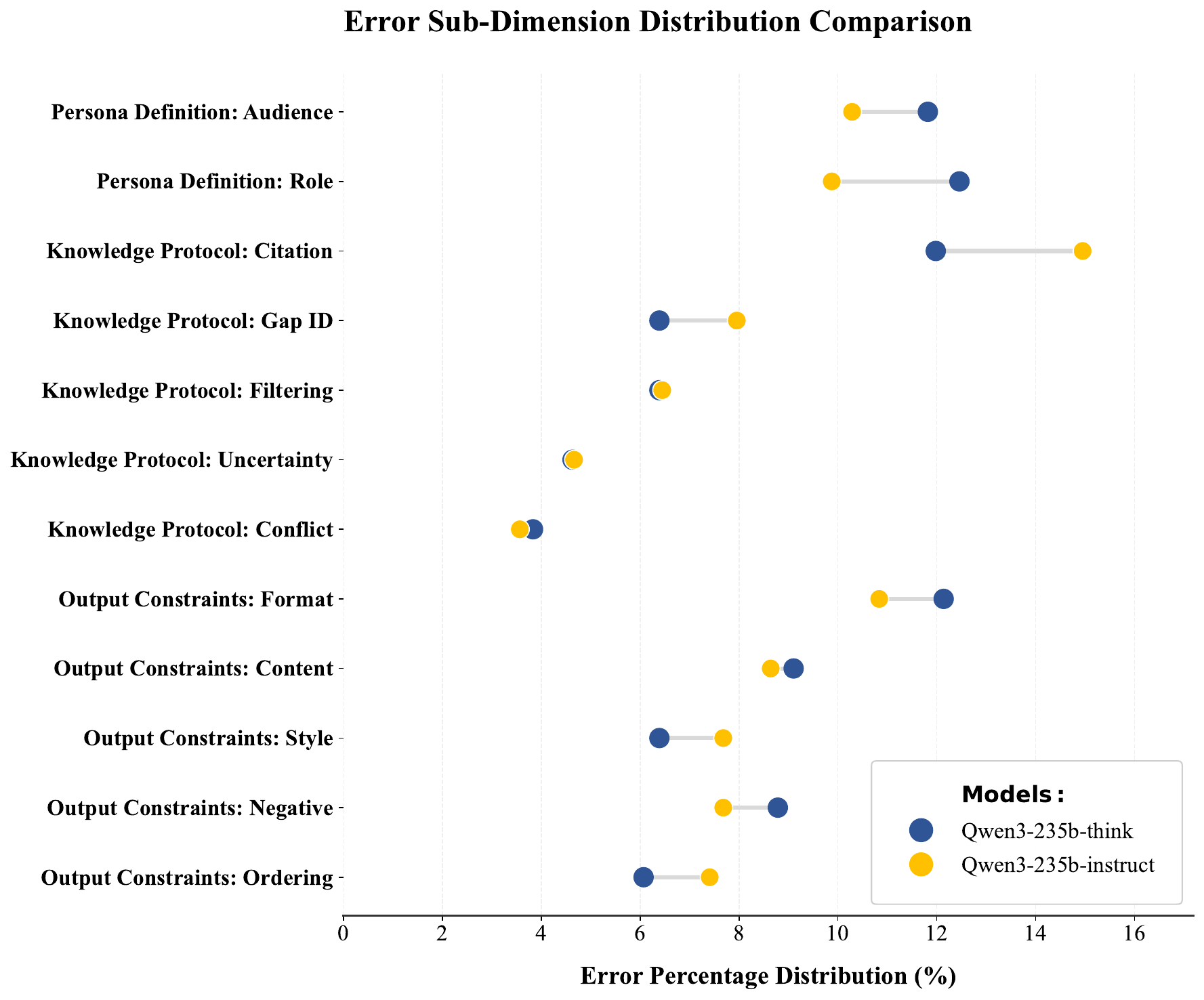}
    \caption{Error distribution by constraint category. Knowledge Interaction Protocols exhibit the highest failure rates, confirming that behavioral judgment, not formatting, is the core bottleneck.}
    \label{fig:subdimension_error}
\end{figure}

\begin{table}[t]
\centering
\footnotesize
\setlength{\tabcolsep}{4pt} 
\begin{tabular}{lc|lc}
\toprule
\textbf{Model} & \textbf{CR} & \textbf{Model} & \textbf{CR} \\
\midrule
DeepSeek-R1$^{\dagger}$ & 44.3 & DeepSeek-V3.1 & 29.8 \\
Gemini-2.5-Pro$^{\dagger}$ & 42.3 & Qwen3-30B-I & 28.5 \\
Qwen3-235B-T$^{\dagger}$ & 41.4 & Qwen3-14B-T$^{\dagger}$ & 27.2 \\
GLM-4.5$^{\dagger}$ & 40.1 & Claude-Opus$^{\dagger}$ & 26.9 \\
Qwen3-235B-I & 37.5 & Claude-Sonnet & 25.9 \\
Qwen3-32B-T$^{\dagger}$ & 36.6 & Qwen3-8B-T$^{\dagger}$ & 23.9 \\
& & GPT-4.1 & 18.5 \\
\bottomrule
\multicolumn{4}{l}{\scriptsize $^{\dagger}$Reasoning-enhanced. CR: Conflict Recog. (\%).}
\end{tabular}
\caption{Conflict recognition rates (CR) across 13 models.}
\label{tab:conflict_recog}
\end{table}

\begin{figure}[t]
    \centering
    \includegraphics[width=\linewidth]{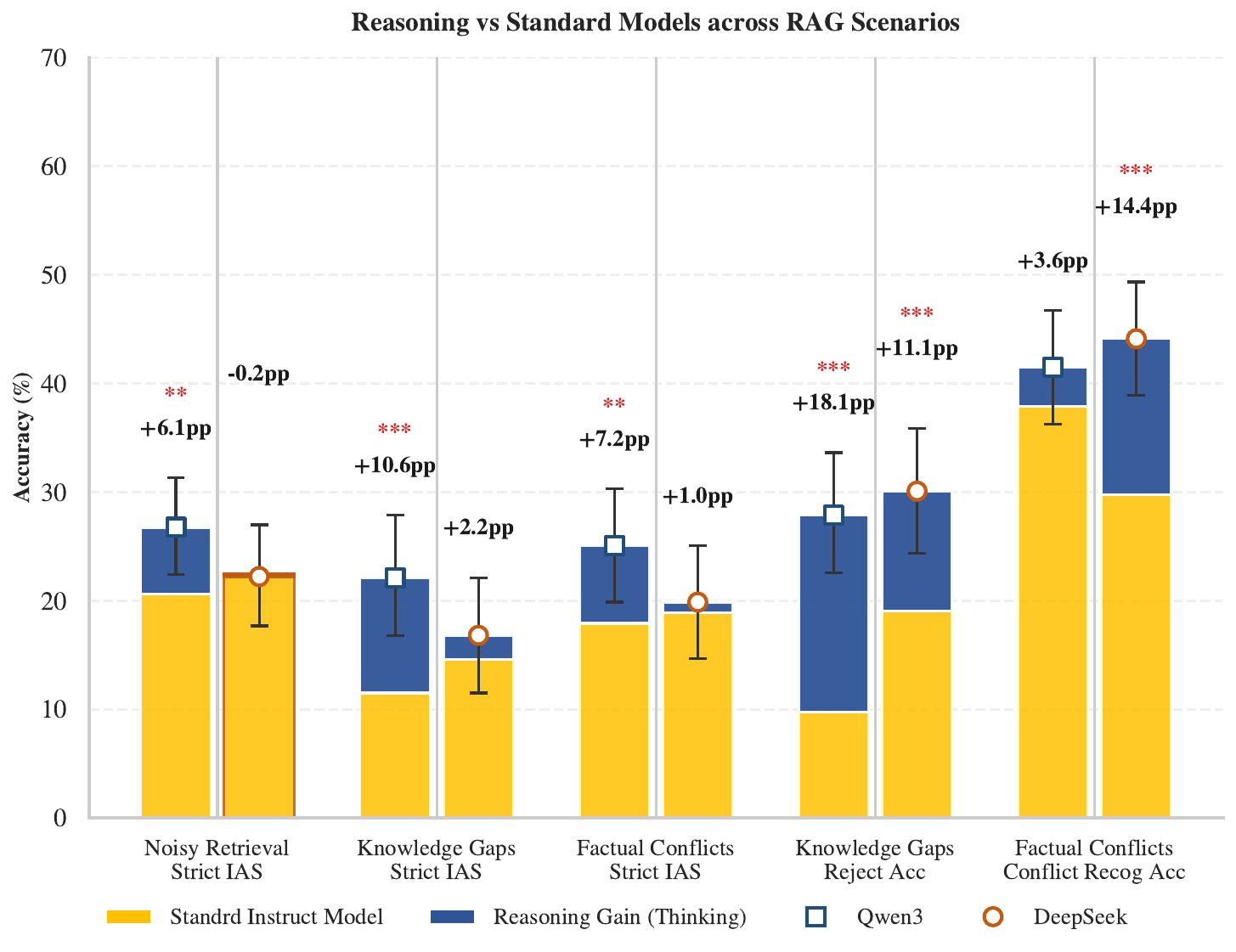}
    \caption{Reasoning vs.\ standard model accuracy across RAG scenarios. Blue segments denote reasoning gains over standard baselines (yellow). Error bars: 95\% CI. $^{*}p<.05$, $^{**}p<.01$, $^{***}p<.001$ (McNemar's test).}
    \label{fig:reason_vs_instruct}
\end{figure}

\begin{figure}[!htb]
    \centering
    \includegraphics[width=\columnwidth]{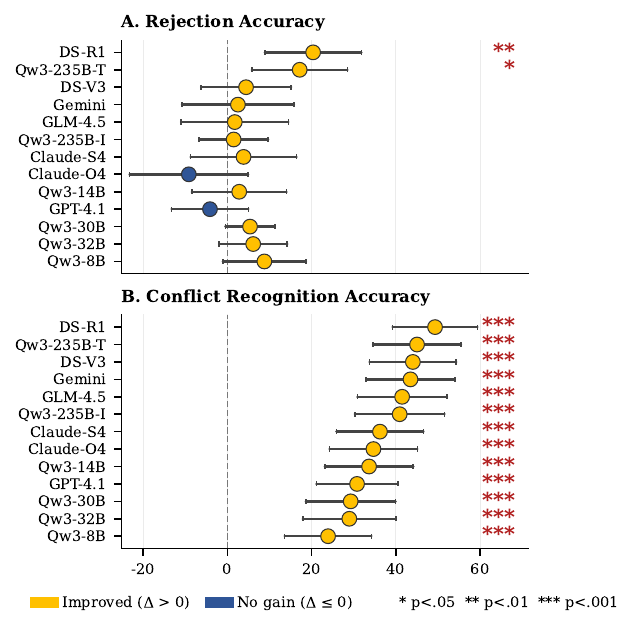}
    \caption{Effect of knowledge interaction protocol on \textbf{(A)} rejection accuracy 
    and \textbf{(B)} conflict recognition. Points show accuracy differences 
    (with/without protocol) with 95\% CIs. 
    The protocol significantly improves conflict recognition in all 13 models, with more modest effects on rejection accuracy 
    (2/13 significant). Independent samples (with/without: 157/69 for A, 113/193 for B); 
    $p$-values from $\chi^2$ test with Yates' correction.}
    \label{fig:robustness_woconstrait_compare}
\end{figure}

\subsection{Main Results}

\paragraph{Finding 1: Orchestration under noisy retrieval.}
Table~\ref{tab:standard-subset} reveals a severe \textbf{adherence collapse}: Loose IAS achieves up to 83.8\%, yet Strict IAS reaches only 26.8\% (Qwen3-235B-Thinking).
This 57-point gap quantifies the \textbf{compositional bottleneck} where models satisfy individual constraints but fail holistic compliance.
Reasoning-enhanced models consistently outperform standard variants, with the largest gains in Knowledge Interaction Protocols.

\paragraph{Finding 2: Rejection under knowledge gaps.}
In production, hallucinating on unanswerable queries is often more harmful than being unhelpful.
Figure~\ref{fig:reject_acc} exposes a pervasive \textbf{helpfulness bias}: Qwen3-30B-Instruct achieves only 6.6\% rejection accuracy, hallucinating in 93.4\% of unanswerable cases.
Reasoning-enhanced models improve substantially (Claude-Opus-4.5: 42.7\%), yet remain far from production-grade reliability.

\paragraph{Finding 3: Conflict recognition.}
Table~\ref{tab:conflict_recog} shows conflict detection remains a bottleneck: top models reach only 40–44\% recognition (DeepSeek-R1: 44.3\%), while GPT-4.1 detects merely 18.5\%.
Figure~\ref{fig:conflict_coverage_corr} shows reasoning-enhanced models achieve a strong positive correlation between recognition and coverage ($\rho{=}+0.90$, $p{<}0.01$), while standard models show no consistent relationship ($\rho{=}-0.50$) with high variance.
This suggests inference-time computation may resolve the traditional safety-informativeness dilemma.
Synthetic and natural conflicts show equivalent difficulty on core metrics (\Cref{subsec:appendix B.2}).

\subsection{Analysis}

\paragraph{Protocol bottleneck.}
Figure~\ref{fig:subdimension_error} decomposes IAS failures by constraint category.
Knowledge Interaction Protocols exhibit the highest error rates and variance, particularly for citation and gap identification.
Comparing Qwen3-235B-Thinking to its Instruct counterpart, the largest reasoning gains occur precisely in these protocol dimensions, confirming that \textbf{judgment under uncertainty} is the core enterprise bottleneck.

\paragraph{Scaling.}
Within Qwen3-Thinking, Strict IAS scales non-linearly (12.3\% at 8B $\rightarrow$ 26.8\% at 235B) while Faithfulness saturates (64.8\% $\rightarrow$ 67.1\%), indicating orchestration is an emergent capability requiring substantial scale.

\paragraph{Reasoning vs.\ standard instruction-tuned variants.}
Figure~\ref{fig:reason_vs_instruct} compares matched reasoning vs.\ standard variants (Qwen3-235B-Thinking vs.\ Qwen3-235B-Instruct; DeepSeek-R1 vs.\ DeepSeek-V3.1).
Across scenarios, reasoning variants exhibit substantial robustness gains: Qwen3-235B-Thinking boosts rejection accuracy by 18.1pp and DeepSeek-R1 improves conflict recognition by 14.4pp, consistent with reduced helpfulness bias.
For Strict IAS, Qwen3-235B-Thinking shows consistent gains (+6.1pp to +10.6pp; $p{<}.01$), while DeepSeek-R1 shows minimal improvement, suggesting architecture-dependent benefits.

\paragraph{Effect of explicit protocols.}
Figure~\ref{fig:robustness_woconstrait_compare} compares instances with explicit Knowledge Interaction Protocol constraints to those without such constraints under the same retrieval failure mode.
Overall, explicit protocols yield a large and consistent gain in conflict recognition across all 13 models, but only modest improvements in rejection under knowledge gaps.
This asymmetry suggests that protocols help most when the failure is explicit in-context (contradictions), whereas proper refusal requires a harder judgment of evidence sufficiency and separating parametric knowledge from retrieved evidence.
Notably, Claude-Opus-4.5 shows a small, non-significant decrease, indicating potential interaction with model-specific safety behaviors.
Because this is an observational comparison (protocol presence is not randomized), we report domain-level breakdowns in \Cref{subsec:appendix C.1}.


\section{Conclusion}
EnterpriseRAG combines complex multi-constraint instructions with three non-ideal retrieval modes across 983 expert-validated instances. Across 13 LLMs, we find a persistent orchestration collapse: even the best model reaches 83.8\% per-constraint adherence (Loose IAS) but only 26.8\% holistic compliance (Strict IAS), leaving a 57-point gap. Robustness failures concentrate in knowledge-interaction protocols: under knowledge gaps, models frequently over-answer despite explicit refusal requirements (with Claude-Opus-4.5 peaking at 42.7\%); under factual conflicts, even the strongest systems recognize contradictions in fewer than half of cases (led by DeepSeek-R1 at 44.3\%).

Practically, our results suggest that enterprise-ready RAG requires (i) training and evaluation targeted at protocol-level judgment (evidence sufficiency, calibrated refusal, and conflict-aware reporting), not just formatting or factuality; and (ii) explicit operational protocols in prompts, which reliably improve conflict handling but are insufficient to solve evidence-gap refusal. EnterpriseRAG provides a realistic and reproducible foundation to measure and close these gaps.

\section{Limitations}
While EnterpriseRAG encompasses six diverse domains, the current scope is limited to text-based RAG. Multimodal contexts, such as those involving charts or images within PDFs, are not yet included. Additionally, our reliance on a reasoning-enhanced LLM as an evaluator, while effective, may introduce bias compared to human evaluation, although our sampling checks indicate high alignment. Finally, the strict adherence metric is binary and stringent; future metrics could explore more nuanced semantic gradations of constraint satisfaction.


\bibliography{bibtex/benchmark,bibtex/model,bibtex/datasets,bibtex/instruction_follow,bibtex/robustness}

@misc{zhao2024retrievalaugmentedgenerationaigeneratedcontent,
      title={Retrieval-Augmented Generation for AI-Generated Content: A Survey}, 
      author={Penghao Zhao and Hailin Zhang and Qinhan Yu and Zhengren Wang and Yunteng Geng and Fangcheng Fu and Ling Yang and Wentao Zhang and Jie Jiang and Bin Cui},
      year={2024},
      eprint={2402.19473},
      archivePrefix={arXiv},
      primaryClass={cs.CV},
      url={https://arxiv.org/abs/2402.19473}, 
}

@misc{gupta2024comprehensivesurveyretrievalaugmentedgeneration,
      title={A Comprehensive Survey of Retrieval-Augmented Generation (RAG): Evolution, Current Landscape and Future Directions}, 
      author={Shailja Gupta and Rajesh Ranjan and Surya Narayan Singh},
      year={2024},
      eprint={2410.12837},
      archivePrefix={arXiv},
      primaryClass={cs.CL},
      url={https://arxiv.org/abs/2410.12837}, 
}

@misc{zhang2025magicmushroomcustomizablebenchmark,
      title={Magic Mushroom: A Customizable Benchmark for Fine-grained Analysis of Retrieval Noise Erosion in RAG Systems}, 
      author={Yuxin Zhang and Yan Wang and Yongrui Chen and Shenyu Zhang and Xinbang Dai and Sheng Bi and Guilin Qi},
      year={2025},
      eprint={2506.03901},
      archivePrefix={arXiv},
      primaryClass={cs.CL},
      url={https://arxiv.org/abs/2506.03901}, 
}

@misc{lee2025magicmultihopgraphbasedbenchmark,
      title={MAGIC: A Multi-Hop and Graph-Based Benchmark for Inter-Context Conflicts in Retrieval-Augmented Generation}, 
      author={Jungyeon Lee and Kangmin Lee and Taeuk Kim},
      year={2025},
      eprint={2507.21544},
      archivePrefix={arXiv},
      primaryClass={cs.CL},
      url={https://arxiv.org/abs/2507.21544}, 
}

@article{xu2025crp,
  title = {{CRP-RAG}: A Retrieval-Augmented Generation Framework for Supporting Complex Logical Reasoning and Knowledge Planning},
  author = {Xu, Kehan and Zhang, Kun and Li, Jingyuan and Huang, Wei and Wang, Yuanzhuo},
  journal = {Electronics},
  volume = {14},
  number = {1},
  pages = {47},
  year = {2025},
  month = {dec},
  publisher = {MDPI},
  doi = {10.3390/electronics14010047},
  url = {https://doi.org/10.3390/electronics14010047}
}

@misc{guo2025llmcentricragmultigranularindexing,
      title={LLM-Centric RAG with Multi-Granular Indexing and Confidence Constraints}, 
      author={Xiaofan Guo and Yaxuan Luan and Yue Kang and Xiangchen Song and Jinxu Guo},
      year={2025},
      eprint={2510.27054},
      archivePrefix={arXiv},
      primaryClass={cs.CL},
      url={https://arxiv.org/abs/2510.27054}, 
}

@inproceedings{choi-etal-2025-conflict,
    title = "Conflict-Aware Soft Prompting for Retrieval-Augmented Generation",
    author = "Choi, Eunseong  and
      Park, June  and
      Lee, Hyeri  and
      Lee, Jongwuk",
    editor = "Christodoulopoulos, Christos  and
      Chakraborty, Tanmoy  and
      Rose, Carolyn  and
      Peng, Violet",
    booktitle = "Proceedings of the 2025 Conference on Empirical Methods in Natural Language Processing",
    month = nov,
    year = "2025",
    address = "Suzhou, China",
    publisher = "Association for Computational Linguistics",
    url = "https://aclanthology.org/2025.emnlp-main.1371/",
    doi = "10.18653/v1/2025.emnlp-main.1371",
    pages = "26969--26983",
    ISBN = "979-8-89176-332-6",
}

@misc{chen2023benchmarkinglargelanguagemodels,
      title={Benchmarking Large Language Models in Retrieval-Augmented Generation}, 
      author={Jiawei Chen and Hongyu Lin and Xianpei Han and Le Sun},
      year={2023},
      eprint={2309.01431},
      archivePrefix={arXiv},
      primaryClass={cs.CL},
      url={https://arxiv.org/abs/2309.01431}, 
}

@inproceedings{yu_ekrag_2025,
	address = {Albuquerque, New Mexico, USA},
	title = {{EKRAG}: {Benchmark} {RAG} for {Enterprise} {Knowledge} {Question} {Answering}},
	isbn = {979-8-89176-229-9},
	shorttitle = {{EKRAG}},
	url = {https://aclanthology.org/2025.knowledgenlp-1.13/},
	doi = {10.18653/v1/2025.knowledgenlp-1.13},
	urldate = {2025-08-07},
	booktitle = {Proceedings of the 4th {International} {Workshop} on {Knowledge}-{Augmented} {Methods} for {Natural} {Language} {Processing}},
	publisher = {Association for Computational Linguistics},
	author = {Yu, Tan and Zhou, Wenfei and Leiyang, Leiyang and Shukla, Aaditya and Mmadugula, Mmadugula and Gundecha, Pritam and Burnett, Nicholas and Xu, Anbang and Viseth, Viseth and Tbar, Tbar and Akkiraju, Rama and Zhang, Vivienne},
	editor = {Shi, Weijia and Yu, Wenhao and Asai, Akari and Jiang, Meng and Durrett, Greg and Hajishirzi, Hannaneh and Zettlemoyer, Luke},
	month = may,
	year = {2025},
	pages = {152--159},
}

@misc{katsis_mtrag_2025,
	title = {{MTRAG}: {A} {Multi}-{Turn} {Conversational} {Benchmark} for {Evaluating} {Retrieval}-{Augmented} {Generation} {Systems}},
	shorttitle = {{MTRAG}},
	url = {http://arxiv.org/abs/2501.03468},
	doi = {10.48550/arXiv.2501.03468},
	urldate = {2025-08-06},
	publisher = {arXiv},
	author = {Katsis, Yannis and Rosenthal, Sara and Fadnis, Kshitij and Gunasekara, Chulaka and Lee, Young-Suk and Popa, Lucian and Shah, Vraj and Zhu, Huaiyu and Contractor, Danish and Danilevsky, Marina},
	month = jan,
	year = {2025},
	note = {arXiv:2501.03468 [cs]},
}

@misc{pipitone_legalbench-rag_2024,
	title = {{LegalBench}-{RAG}: {A} {Benchmark} for {Retrieval}-{Augmented} {Generation} in the {Legal} {Domain}},
	shorttitle = {{LegalBench}-{RAG}},
	url = {http://arxiv.org/abs/2408.10343},
	doi = {10.48550/arXiv.2408.10343},
	urldate = {2025-08-06},
	publisher = {arXiv},
	author = {Pipitone, Nicholas and Alami, Ghita Houir},
	month = aug,
	year = {2024},
	note = {arXiv:2408.10343 [cs]},
}

@misc{zhu_rageval_2025,
	title = {{RAGEval}: {Scenario} {Specific} {RAG} {Evaluation} {Dataset} {Generation} {Framework}},
	shorttitle = {{RAGEval}},
	url = {http://arxiv.org/abs/2408.01262},
	doi = {10.48550/arXiv.2408.01262},
	urldate = {2025-08-06},
	publisher = {arXiv},
	author = {Zhu, Kunlun and Luo, Yifan and Xu, Dingling and Yan, Yukun and Liu, Zhenghao and Yu, Shi and Wang, Ruobing and Wang, Shuo and Li, Yishan and Zhang, Nan and Han, Xu and Liu, Zhiyuan and Sun, Maosong},
	month = mar,
	year = {2025},
	note = {arXiv:2408.01262 [cs]},
}

@misc{lyu_crud-rag_2024,
	title = {{CRUD}-{RAG}: {A} {Comprehensive} {Chinese} {Benchmark} for {Retrieval}-{Augmented} {Generation} of {Large} {Language} {Models}},
	shorttitle = {{CRUD}-{RAG}},
	url = {http://arxiv.org/abs/2401.17043},
	doi = {10.48550/arXiv.2401.17043},
	urldate = {2025-08-06},
	publisher = {arXiv},
	author = {Lyu, Yuanjie and Li, Zhiyu and Niu, Simin and Xiong, Feiyu and Tang, Bo and Wang, Wenjin and Wu, Hao and Liu, Huanyong and Xu, Tong and Chen, Enhong},
	month = jul,
	year = {2024},
	note = {arXiv:2401.17043 [cs]},
}

@misc{friel_ragbench_2025,
	title = {{RAGBench}: {Explainable} {Benchmark} for {Retrieval}-{Augmented} {Generation} {Systems}},
	shorttitle = {{RAGBench}},
	url = {http://arxiv.org/abs/2407.11005},
	doi = {10.48550/arXiv.2407.11005},
	urldate = {2025-08-06},
	publisher = {arXiv},
	author = {Friel, Robert and Belyi, Masha and Sanyal, Atindriyo},
	month = jan,
	year = {2025},
	note = {arXiv:2407.11005 [cs]},
}

@inproceedings{jin_pubmedqa_2019,
	address = {Hong Kong, China},
	title = {{PubMedQA}: {A} {Dataset} for {Biomedical} {Research} {Question} {Answering}},
	shorttitle = {{PubMedQA}},
	url = {https://www.aclweb.org/anthology/D19-1259},
	doi = {10.18653/v1/D19-1259},
	language = {en},
	urldate = {2025-08-06},
	booktitle = {Proceedings of the 2019 {Conference} on {Empirical} {Methods} in {Natural} {Language} {Processing} and the 9th {International} {Joint} {Conference} on {Natural} {Language} {Processing} ({EMNLP}-{IJCNLP})},
	publisher = {Association for Computational Linguistics},
	author = {Jin, Qiao and Dhingra, Bhuwan and Liu, Zhengping and Cohen, William and Lu, Xinghua},
	year = {2019},
	pages = {2567--2577},
}

@misc{sorodoc_garage_2025,
	title = {{GaRAGe}: {A} {Benchmark} with {Grounding} {Annotations} for {RAG} {Evaluation}},
	shorttitle = {{GaRAGe}},
	url = {http://arxiv.org/abs/2506.07671},
	doi = {10.48550/arXiv.2506.07671},
	urldate = {2025-11-04},
	publisher = {arXiv},
	author = {Sorodoc, Ionut-Teodor and Ribeiro, Leonardo F. R. and Blloshmi, Rexhina and Davis, Christopher and Gispert, Adrià de},
	month = jun,
	year = {2025},
	note = {arXiv:2506.07671 [cs]},
}

@misc{dong_toward_2024,
	title = {Toward {General} {Instruction}-{Following} {Alignment} for {Retrieval}-{Augmented} {Generation}},
	url = {http://arxiv.org/abs/2410.09584},
	doi = {10.48550/arXiv.2410.09584},
	urldate = {2025-11-18},
	publisher = {arXiv},
	author = {Dong, Guanting and Song, Xiaoshuai and Zhu, Yutao and Qiao, Runqi and Dou, Zhicheng and Wen, Ji-Rong},
	month = oct,
	year = {2024},
	note = {arXiv:2410.09584 [cs]},
}

@misc{yang_crag_2024,
	title = {{CRAG} -- {Comprehensive} {RAG} {Benchmark}},
	url = {http://arxiv.org/abs/2406.04744},
	doi = {10.48550/arXiv.2406.04744},
	urldate = {2025-11-18},
	publisher = {arXiv},
	author = {Yang, Xiao and Sun, Kai and Xin, Hao and Sun, Yushi and Bhalla, Nikita and Chen, Xiangsen and Choudhary, Sajal and Gui, Rongze Daniel and Jiang, Ziran Will and Jiang, Ziyu and Kong, Lingkun and Moran, Brian and Wang, Jiaqi and Xu, Yifan Ethan and Yan, An and Yang, Chenyu and Yuan, Eting and Zha, Hanwen and Tang, Nan and Chen, Lei and Scheffer, Nicolas and Liu, Yue and Shah, Nirav and Wanga, Rakesh and Kumar, Anuj and Yih, Wen-tau and Dong, Xin Luna},
	month = nov,
	year = {2024},
	note = {arXiv:2406.04744 [cs]},
}

@article{chen_disc-finllm_nodate,
	title = {{DISC}-{FinLLM}: {A} {Chinese} {Financial} {Large} {Language} {Model} based on {Multiple} {Experts} {Fine}-tuning},
	language = {en},
 year = {2023},
	author = {Chen, Wei and Wang, Qiushi and Long, Zefei and Zhang, Xianyin and Lu, Zhongtian and Li, Bingxuan and Wang, Siyuan and Xu, Jiarong and Bai, Xiang and Huang, Xuanjing and Wei, Zhongyu},
}

@misc{gao_enabling_2023,
	title = {Enabling {Large} {Language} {Models} to {Generate} {Text} with {Citations}},
	url = {http://arxiv.org/abs/2305.14627},
	doi = {10.48550/arXiv.2305.14627},
	urldate = {2025-12-02},
	publisher = {arXiv},
	author = {Gao, Tianyu and Yen, Howard and Yu, Jiatong and Chen, Danqi},
	month = oct,
	year = {2023},
	note = {arXiv:2305.14627 [cs]},
}

@misc{es_ragas_2025,
	title = {Ragas: {Automated} {Evaluation} of {Retrieval} {Augmented} {Generation}},
	shorttitle = {Ragas},
	url = {http://arxiv.org/abs/2309.15217},
	doi = {10.48550/arXiv.2309.15217},
	urldate = {2025-12-04},
	publisher = {arXiv},
	author = {Es, Shahul and James, Jithin and Espinosa-Anke, Luis and Schockaert, Steven},
	month = apr,
	year = {2025},
	note = {arXiv:2309.15217 [cs]},
}

@misc{tang_multihop-rag_2024,
	title = {{MultiHop}-{RAG}: {Benchmarking} {Retrieval}-{Augmented} {Generation} for {Multi}-{Hop} {Queries}},
	shorttitle = {{MultiHop}-{RAG}},
	url = {http://arxiv.org/abs/2401.15391},
	doi = {10.48550/arXiv.2401.15391},
	urldate = {2025-12-18},
	publisher = {arXiv},
	author = {Tang, Yixuan and Yang, Yi},
	month = jan,
	year = {2024},
	note = {arXiv:2401.15391 [cs]},
}

@misc{yang2018hotpotqadatasetdiverseexplainable,
      title={HotpotQA: A Dataset for Diverse, Explainable Multi-hop Question Answering}, 
      author={Zhilin Yang and Peng Qi and Saizheng Zhang and Yoshua Bengio and William W. Cohen and Ruslan Salakhutdinov and Christopher D. Manning},
      year={2018},
      eprint={1809.09600},
      archivePrefix={arXiv},
      primaryClass={cs.CL},
      url={https://arxiv.org/abs/1809.09600}, 
}

@misc{FinLongEval,
  author  = {Xinguang Jiang and Sihan Hu and Dingfu Yu and Yuhao Zhang and Zhongliang Yang and Yu Li and Linna Zhou and {Valuesimplex AI Lab}},
  title   = {{FinLongEval}},
  howpublished = {\url{https://github.com/valuesimplex/FinLongEval}},
  year    = {2023},
  month   = dec
}

@inproceedings{jiang-etal-2024-followbench,
    title = "{F}ollow{B}ench: A Multi-level Fine-grained Constraints Following Benchmark for Large Language Models",
    author = "Jiang, Yuxin  and
      Wang, Yufei  and
      Zeng, Xingshan  and
      Zhong, Wanjun  and
      Li, Liangyou  and
      Mi, Fei  and
      Shang, Lifeng  and
      Jiang, Xin  and
      Liu, Qun  and
      Wang, Wei",
    editor = "Ku, Lun-Wei  and
      Martins, Andre  and
      Srikumar, Vivek",
    booktitle = "Proceedings of the 62nd Annual Meeting of the Association for Computational Linguistics (Volume 1: Long Papers)",
    month = aug,
    year = "2024",
    address = "Bangkok, Thailand",
    publisher = "Association for Computational Linguistics",
    url = "https://aclanthology.org/2024.acl-long.257/",
    doi = "10.18653/v1/2024.acl-long.257",
    pages = "4667--4688"
}

@misc{zhou_instruction-following_2023,
	title = {Instruction-Following Evaluation for Large Language Models},
	url = {http://arxiv.org/abs/2311.07911},
	doi = {10.48550/arXiv.2311.07911},
	number = {{arXiv}:2311.07911},
	publisher = {{arXiv}},
	author = {Zhou, Jeffrey and Lu, Tianjian and Mishra, Swaroop and Brahma, Siddhartha and Basu, Sujoy and Luan, Yi and Zhou, Denny and Hou, Le},
	urldate = {2025-12-05},
	date = {2023-11-14},
    year = {2023},
	eprinttype = {arxiv},
	eprint = {2311.07911 [cs]},
}

@misc{qin_sysbench_2024,
	title = {{SysBench}: Can Large Language Models Follow System Messages?},
	url = {http://arxiv.org/abs/2408.10943},
	doi = {10.48550/arXiv.2408.10943},
	shorttitle = {{SysBench}},
	number = {{arXiv}:2408.10943},
	publisher = {{arXiv}},
	author = {Qin, Yanzhao and Zhang, Tao and Zhang, Tao and Shen, Yanjun and Luo, Wenjing and Sun, Haoze and Zhang, Yan and Qiao, Yujing and Chen, Weipeng and Zhou, Zenan and Zhang, Wentao and Cui, Bin},
	urldate = {2025-12-05},
	date = {2024-08-20},
    year = {2024},
	eprinttype = {arxiv},
	eprint = {2408.10943 [cs]},
	note = {version: 1},
}

@book{wen_benchmarking_2024,
	title = {Benchmarking Complex Instruction-Following with Multiple Constraints Composition},
	author = {Wen, Bosi and Ke, Pei and Gu, Xiaotao and Wu, Lindong and Huang, Hao and Zhou, Jinfeng and Li, Wenchuang and Hu, Binxin and Gao, Wendy and Xu, Jiaxin and Liu, Yiming and Tang, Jie and Wang, Hongning and Huang, Minlie},
	date = {2024-07-04},
    year = {2024},
	doi = {10.48550/arXiv.2407.03978},
}

@inproceedings{zou_eifbench_2025,
	location = {Suzhou, China},
	title = {{EIFBENCH}: Extremely Complex Instruction Following Benchmark for Large Language Models},
	isbn = {979-8-89176-332-6},
	url = {https://aclanthology.org/2025.emnlp-main.1059/},
	doi = {10.18653/v1/2025.emnlp-main.1059},
	shorttitle = {{EIFBENCH}},
	eventtitle = {{EMNLP} 2025},
	pages = {20941--20964},
	booktitle = {Proceedings of the 2025 Conference on Empirical Methods in Natural Language Processing},
	publisher = {Association for Computational Linguistics},
	author = {Zou, Tao and Zhang, Xinghua and Yu, Haiyang and Wang, Minzheng and Huang, Fei and Li, Yongbin},
	editor = {Christodoulopoulos, Christos and Chakraborty, Tanmoy and Rose, Carolyn and Peng, Violet},
	urldate = {2025-12-16},
	date = {2025-11},
    year = {2025},
}

@misc{yang_qwen3_2025,
	title = {Qwen3 Technical Report},
	url = {http://arxiv.org/abs/2505.09388},
	doi = {10.48550/arXiv.2505.09388},
	number = {{arXiv}:2505.09388},
	publisher = {{arXiv}},
	author = {Yang, An and Li, Anfeng and Yang, Baosong and Zhang, Beichen and Hui, Binyuan and Zheng, Bo and Yu, Bowen and Gao, Chang and Huang, Chengen and Lv, Chenxu and Zheng, Chujie and Liu, Dayiheng and Zhou, Fan and Huang, Fei and Hu, Feng and Ge, Hao and Wei, Haoran and Lin, Huan and Tang, Jialong and Yang, Jian and Tu, Jianhong and Zhang, Jianwei and Yang, Jianxin and Yang, Jiaxi and Zhou, Jing and Zhou, Jingren and Lin, Junyang and Dang, Kai and Bao, Keqin and Yang, Kexin and Yu, Le and Deng, Lianghao and Li, Mei and Xue, Mingfeng and Li, Mingze and Zhang, Pei and Wang, Peng and Zhu, Qin and Men, Rui and Gao, Ruize and Liu, Shixuan and Luo, Shuang and Li, Tianhao and Tang, Tianyi and Yin, Wenbiao and Ren, Xingzhang and Wang, Xinyu and Zhang, Xinyu and Ren, Xuancheng and Fan, Yang and Su, Yang and Zhang, Yichang and Zhang, Yinger and Wan, Yu and Liu, Yuqiong and Wang, Zekun and Cui, Zeyu and Zhang, Zhenru and Zhou, Zhipeng and Qiu, Zihan},
	urldate = {2025-12-16},
	date = {2025-05-14},
 year = {2025}, 
	eprinttype = {arxiv},
	eprint = {2505.09388 [cs]},
}

@misc{deepseek-ai_deepseek-r1_2025,
	title = {{DeepSeek}-R1: Incentivizing Reasoning Capability in {LLMs} via Reinforcement Learning},
	url = {http://arxiv.org/abs/2501.12948},
	doi = {10.48550/arXiv.2501.12948},
	shorttitle = {{DeepSeek}-R1},
	number = {{arXiv}:2501.12948},
	publisher = {{arXiv}},
	author = {{DeepSeek}-{AI} and Guo, Daya and Yang, Dejian and Zhang, Haowei and Song, Junxiao and Zhang, Ruoyu and Xu, Runxin and Zhu, Qihao and Ma, Shirong and Wang, Peiyi and Bi, Xiao and Zhang, Xiaokang and Yu, Xingkai and Wu, Yu and Wu, Z. F. and Gou, Zhibin and Shao, Zhihong and Li, Zhuoshu and Gao, Ziyi and Liu, Aixin and Xue, Bing and Wang, Bingxuan and Wu, Bochao and Feng, Bei and Lu, Chengda and Zhao, Chenggang and Deng, Chengqi and Zhang, Chenyu and Ruan, Chong and Dai, Damai and Chen, Deli and Ji, Dongjie and Li, Erhang and Lin, Fangyun and Dai, Fucong and Luo, Fuli and Hao, Guangbo and Chen, Guanting and Li, Guowei and Zhang, H. and Bao, Han and Xu, Hanwei and Wang, Haocheng and Ding, Honghui and Xin, Huajian and Gao, Huazuo and Qu, Hui and Li, Hui and Guo, Jianzhong and Li, Jiashi and Wang, Jiawei and Chen, Jingchang and Yuan, Jingyang and Qiu, Junjie and Li, Junlong and Cai, J. L. and Ni, Jiaqi and Liang, Jian and Chen, Jin and Dong, Kai and Hu, Kai and Gao, Kaige and Guan, Kang and Huang, Kexin and Yu, Kuai and Wang, Lean and Zhang, Lecong and Zhao, Liang and Wang, Litong and Zhang, Liyue and Xu, Lei and Xia, Leyi and Zhang, Mingchuan and Zhang, Minghua and Tang, Minghui and Li, Meng and Wang, Miaojun and Li, Mingming and Tian, Ning and Huang, Panpan and Zhang, Peng and Wang, Qiancheng and Chen, Qinyu and Du, Qiushi and Ge, Ruiqi and Zhang, Ruisong and Pan, Ruizhe and Wang, Runji and Chen, R. J. and Jin, R. L. and Chen, Ruyi and Lu, Shanghao and Zhou, Shangyan and Chen, Shanhuang and Ye, Shengfeng and Wang, Shiyu and Yu, Shuiping and Zhou, Shunfeng and Pan, Shuting and Li, S. S. and Zhou, Shuang and Wu, Shaoqing and Ye, Shengfeng and Yun, Tao and Pei, Tian and Sun, Tianyu and Wang, T. and Zeng, Wangding and Zhao, Wanjia and Liu, Wen and Liang, Wenfeng and Gao, Wenjun and Yu, Wenqin and Zhang, Wentao and Xiao, W. L. and An, Wei and Liu, Xiaodong and Wang, Xiaohan and Chen, Xiaokang and Nie, Xiaotao and Cheng, Xin and Liu, Xin and Xie, Xin and Liu, Xingchao and Yang, Xinyu and Li, Xinyuan and Su, Xuecheng and Lin, Xuheng and Li, X. Q. and Jin, Xiangyue and Shen, Xiaojin and Chen, Xiaosha and Sun, Xiaowen and Wang, Xiaoxiang and Song, Xinnan and Zhou, Xinyi and Wang, Xianzu and Shan, Xinxia and Li, Y. K. and Wang, Y. Q. and Wei, Y. X. and Zhang, Yang and Xu, Yanhong and Li, Yao and Zhao, Yao and Sun, Yaofeng and Wang, Yaohui and Yu, Yi and Zhang, Yichao and Shi, Yifan and Xiong, Yiliang and He, Ying and Piao, Yishi and Wang, Yisong and Tan, Yixuan and Ma, Yiyang and Liu, Yiyuan and Guo, Yongqiang and Ou, Yuan and Wang, Yuduan and Gong, Yue and Zou, Yuheng and He, Yujia and Xiong, Yunfan and Luo, Yuxiang and You, Yuxiang and Liu, Yuxuan and Zhou, Yuyang and Zhu, Y. X. and Xu, Yanhong and Huang, Yanping and Li, Yaohui and Zheng, Yi and Zhu, Yuchen and Ma, Yunxian and Tang, Ying and Zha, Yukun and Yan, Yuting and Ren, Z. Z. and Ren, Zehui and Sha, Zhangli and Fu, Zhe and Xu, Zhean and Xie, Zhenda and Zhang, Zhengyan and Hao, Zhewen and Ma, Zhicheng and Yan, Zhigang and Wu, Zhiyu and Gu, Zihui and Zhu, Zijia and Liu, Zijun and Li, Zilin and Xie, Ziwei and Song, Ziyang and Pan, Zizheng and Huang, Zhen and Xu, Zhipeng and Zhang, Zhongyu and Zhang, Zhen},
	urldate = {2025-12-16},
	date = {2025-01-22},
 year = {2025}, 
	eprinttype = {arxiv},
	eprint = {2501.12948 [cs]},
}

@misc{team_glm-45_2025,
	title = {{GLM}-4.5: Agentic, Reasoning, and Coding ({ARC}) Foundation Models},
	url = {http://arxiv.org/abs/2508.06471},
	doi = {10.48550/arXiv.2508.06471},
	shorttitle = {{GLM}-4.5},
	number = {{arXiv}:2508.06471},
	publisher = {{arXiv}},
	author = {Team, {GLM}-4 5 and Zeng, Aohan and Lv, Xin and Zheng, Qinkai and Hou, Zhenyu and Chen, Bin and Xie, Chengxing and Wang, Cunxiang and Yin, Da and Zeng, Hao and Zhang, Jiajie and Wang, Kedong and Zhong, Lucen and Liu, Mingdao and Lu, Rui and Cao, Shulin and Zhang, Xiaohan and Huang, Xuancheng and Wei, Yao and Cheng, Yean and An, Yifan and Niu, Yilin and Wen, Yuanhao and Bai, Yushi and Du, Zhengxiao and Wang, Zihan and Zhu, Zilin and Zhang, Bohan and Wen, Bosi and Wu, Bowen and Xu, Bowen and Huang, Can and Zhao, Casey and Cai, Changpeng and Yu, Chao and Li, Chen and Ge, Chendi and Huang, Chenghua and Zhang, Chenhui and Xu, Chenxi and Zhu, Chenzheng and Li, Chuang and Yin, Congfeng and Lin, Daoyan and Yang, Dayong and Jiang, Dazhi and Ai, Ding and Zhu, Erle and Wang, Fei and Pan, Gengzheng and Wang, Guo and Sun, Hailong and Li, Haitao and Li, Haiyang and Hu, Haiyi and Zhang, Hanyu and Peng, Hao and Tai, Hao and Zhang, Haoke and Wang, Haoran and Yang, Haoyu and Liu, He and Zhao, He and Liu, Hongwei and Yan, Hongxi and Liu, Huan and Chen, Huilong and Li, Ji and Zhao, Jiajing and Ren, Jiamin and Jiao, Jian and Zhao, Jiani and Yan, Jianyang and Wang, Jiaqi and Gui, Jiayi and Zhao, Jiayue and Liu, Jie and Li, Jijie and Li, Jing and Lu, Jing and Wang, Jingsen and Yuan, Jingwei and Li, Jingxuan and Du, Jingzhao and Du, Jinhua and Liu, Jinxin and Zhi, Junkai and Gao, Junli and Wang, Ke and Yang, Lekang and Xu, Liang and Fan, Lin and Wu, Lindong and Ding, Lintao and Wang, Lu and Zhang, Man and Li, Minghao and Xu, Minghuan and Zhao, Mingming and Zhai, Mingshu and Du, Pengfan and Dong, Qian and Lei, Shangde and Tu, Shangqing and Yang, Shangtong and Lu, Shaoyou and Li, Shijie and Li, Shuang and Shuang-Li and Yang, Shuxun and Yi, Sibo and Yu, Tianshu and Tian, Wei and Wang, Weihan and Yu, Wenbo and Tam, Weng Lam and Liang, Wenjie and Liu, Wentao and Wang, Xiao and Jia, Xiaohan and Gu, Xiaotao and Ling, Xiaoying and Wang, Xin and Fan, Xing and Pan, Xingru and Zhang, Xinyuan and Zhang, Xinze and Fu, Xiuqing and Zhang, Xunkai and Xu, Yabo and Wu, Yandong and Lu, Yida and Wang, Yidong and Zhou, Yilin and Pan, Yiming and Zhang, Ying and Wang, Yingli and Li, Yingru and Su, Yinpei and Geng, Yipeng and Zhu, Yitong and Yang, Yongkun and Li, Yuhang and Wu, Yuhao and Li, Yujiang and Liu, Yunan and Wang, Yunqing and Li, Yuntao and Zhang, Yuxuan and Liu, Zezhen and Yang, Zhen and Zhou, Zhengda and Qiao, Zhongpei and Feng, Zhuoer and Liu, Zhuorui and Zhang, Zichen and Wang, Zihan and Yao, Zijun and Wang, Zikang and Liu, Ziqiang and Chai, Ziwei and Li, Zixuan and Zhao, Zuodong and Chen, Wenguang and Zhai, Jidong and Xu, Bin and Huang, Minlie and Wang, Hongning and Li, Juanzi and Dong, Yuxiao and Tang, Jie},
	urldate = {2025-12-16},
	date = {2025-08-08},
 year = {2025}, 
	eprinttype = {arxiv},
	eprint = {2508.06471 [cs]},
}

@misc{comanici_gemini_2025,
    title = {Gemini 2.5: Pushing the Frontier with Advanced Reasoning, Multimodality, Long Context, and Next Generation Agentic Capabilities},
    url = {http://arxiv.org/abs/2507.06261},
    doi = {10.48550/arXiv.2507.06261},
    shorttitle = {Gemini 2.5},
    number = {{arXiv}:2507.06261},
    publisher = {{arXiv}},
    author = {Comanici, Gheorghe and Bieber, Eric and Schaekermann, Mike and Pasupat, Ice and Sachdeva, Noveen and Dhillon, Inderjit and Blistein, Marcel and Ram, Ori and others},
    urldate = {2025-12-16},
    date = {2025-10-16},
    year = {2025},
    eprinttype = {arxiv},
    eprint = {2507.06261}
}

@misc{team_kimi_2025,
	title = {Kimi K2: Open Agentic Intelligence},
	url = {http://arxiv.org/abs/2507.20534},
	doi = {10.48550/arXiv.2507.20534},
	shorttitle = {Kimi K2},
	number = {{arXiv}:2507.20534},
	publisher = {{arXiv}},
	author = {Team, Kimi and Bai, Yifan and Bao, Yiping and Chen, Guanduo and Chen, Jiahao and Chen, Ningxin and Chen, Ruijue and Chen, Yanru and Chen, Yuankun and Chen, Yutian and Chen, Zhuofu and Cui, Jialei and Ding, Hao and Dong, Mengnan and Du, Angang and Du, Chenzhuang and Du, Dikang and Du, Yulun and Fan, Yu and Feng, Yichen and Fu, Kelin and Gao, Bofei and Gao, Hongcheng and Gao, Peizhong and Gao, Tong and Gu, Xinran and Guan, Longyu and Guo, Haiqing and Guo, Jianhang and Hu, Hao and Hao, Xiaoru and He, Tianhong and He, Weiran and He, Wenyang and Hong, Chao and Hu, Yangyang and Hu, Zhenxing and Huang, Weixiao and Huang, Zhiqi and Huang, Zihao and Jiang, Tao and Jiang, Zhejun and Jin, Xinyi and Kang, Yongsheng and Lai, Guokun and Li, Cheng and Li, Fang and Li, Haoyang and Li, Ming and Li, Wentao and Li, Yanhao and Li, Yiwei and Li, Zhaowei and Li, Zheming and Lin, Hongzhan and Lin, Xiaohan and Lin, Zongyu and Liu, Chengyin and Liu, Chenyu and Liu, Hongzhang and Liu, Jingyuan and Liu, Junqi and Liu, Liang and Liu, Shaowei and Liu, T. Y. and Liu, Tianwei and Liu, Weizhou and Liu, Yangyang and Liu, Yibo and Liu, Yiping and Liu, Yue and Liu, Zhengying and Lu, Enzhe and Lu, Lijun and Ma, Shengling and Ma, Xinyu and Ma, Yingwei and Mao, Shaoguang and Mei, Jie and Men, Xin and Miao, Yibo and Pan, Siyuan and Peng, Yebo and Qin, Ruoyu and Qu, Bowen and Shang, Zeyu and Shi, Lidong and Shi, Shengyuan and Song, Feifan and Su, Jianlin and Su, Zhengyuan and Sun, Xinjie and Sung, Flood and Tang, Heyi and Tao, Jiawen and Teng, Qifeng and Wang, Chensi and Wang, Dinglu and Wang, Feng and Wang, Haiming and Wang, Jianzhou and Wang, Jiaxing and Wang, Jinhong and Wang, Shengjie and Wang, Shuyi and Wang, Yao and Wang, Yejie and Wang, Yiqin and Wang, Yuxin and Wang, Yuzhi and Wang, Zhaoji and Wang, Zhengtao and Wang, Zhexu and Wei, Chu and Wei, Qianqian and Wu, Wenhao and Wu, Xingzhe and Wu, Yuxin and Xiao, Chenjun and Xie, Xiaotong and Xiong, Weimin and Xu, Boyu and Xu, Jing and Xu, Jinjing and Xu, L. H. and Xu, Lin and Xu, Suting and Xu, Weixin and Xu, Xinran and Xu, Yangchuan and Xu, Ziyao and Yan, Junjie and Yan, Yuzi and Yang, Xiaofei and Yang, Ying and Yang, Zhen and Yang, Zhilin and Yang, Zonghan and Yao, Haotian and Yao, Xingcheng and Ye, Wenjie and Ye, Zhuorui and Yin, Bohong and Yu, Longhui and Yuan, Enming and Yuan, Hongbang and Yuan, Mengjie and Zhan, Haobing and Zhang, Dehao and Zhang, Hao and Zhang, Wanlu and Zhang, Xiaobin and Zhang, Yangkun and Zhang, Yizhi and Zhang, Yongting and Zhang, Yu and Zhang, Yutao and Zhang, Yutong and Zhang, Zheng and Zhao, Haotian and Zhao, Yikai and Zheng, Huabin and Zheng, Shaojie and Zhou, Jianren and Zhou, Xinyu and Zhou, Zaida and Zhu, Zhen and Zhuang, Weiyu and Zu, Xinxing},
	urldate = {2025-12-16},
	date = {2025-07-28},
   year = {2025}, 
	eprinttype = {arxiv},
	eprint = {2507.20534 [cs]},
}

@online{Claude_Opus_4_5_model_card,
  author       = {{Anthropic}},
  title        = {Introducing {Claude} {Opus} 4.5},
  url          = {https://www.anthropic.com/news/claude-opus-4-5},
  urldate      = {2025-12-18},
  year         = {2025},
  note         = {Blog post}
}

@online{gpt_4_1,
  author       = {{OpenAI}},
  title        = {Introducing {GPT}-4.1 in the {API}},
  url          = {https://openai.com/index/gpt-4-1/},
  urldate      = {2025-04-14},
  year         = {2025},
  note         = {Blog post}
}

@misc{zeng_rare_2025,
    title = {{RARE}: Retrieval-Aware Robustness Evaluation for Retrieval-Augmented Generation Systems},
    author = {Zeng, Yixiao and Cao, Tianyu and Wang, Danqing and Zhao, Xinran and Qiu, Zimeng and Ziyadi, Morteza and Wu, Tongshuang and Li, Lei},
    year = {2025},
    month = {10},
    eprint = {2506.00789},
    archivePrefix = {arXiv},
    primaryClass = {cs.CL},
    url = {https://arxiv.org/abs/2506.00789},
    doi = {10.48550/arXiv.2506.00789},
}

\appendix

\section{Dataset Details \& Statistics}
\label{sec:appendix A}

\subsection{Data Source, Privacy, and Domains}
\label{subsec:appendix A.1}

Our benchmark is constructed from real-world enterprise operational logs in China, collected under internal data usage agreements. The native language of EnterpriseRAG is Chinese. To ensure privacy, all raw logs undergo a strict multi-stage desensitization pipeline, including rule-based removal of sensitive fields and manual expert review. Consequently, no personally identifiable information (PII), proprietary identifiers, or confidential business content is included.

Due to privacy and compliance constraints, the raw data cannot be publicly released. However, we release the desensitized benchmark dataset, along with a reproducible synthetic data generation pipeline and a representative sample subset, to enable independent verification and follow-up research.

\paragraph{Domain Specifications.}
Based on the aforementioned data sources, we select six vertical domains constructed according to three complementary criteria: (1) \textit{Industrial Prevalence}—domains with substantial enterprise RAG deployments; (2) \textit{Data Accessibility}—availability of production logs and domain expertise; and (3) \textit{Task Diversity}—coverage of distinct knowledge processing paradigms.

\begin{itemize}[leftmargin=*,nosep]
    \item \textbf{Energy}: Procedural QA with version drift (e.g., equipment maintenance protocols across ERP system transitions).
    \item \textbf{Medical}: Clinical abstraction from fragmented dialogues, including discharge summaries and treatment synthesis.
    \item \textbf{Legal}: Multi-hop statute-case correlation with jurisdictional hierarchies.
    \item \textbf{Financial}: Investment advisory and policy interpretation tasks, partially leveraging FinLongEval~\citep{FinLongEval}.
    \item \textbf{Party Building}: Regulatory knowledge management and policy interpretation in organizational contexts.
    \item \textbf{Web Search}: Real-time queries emphasizing information freshness and source credibility.
\end{itemize}

\subsection{Data construction process}
\label{subsec:appendix A.2}

Our data construction follows a four-stage pipeline: (1) \textbf{Query Collection}: We gather 491 authentic user queries from operational logs across six domains, ensuring diversity in information needs and complexity levels. (2) \textbf{Instruction Synthesis}: For each query, we apply the constraint fusion protocol (Section 3.3.2) to generate complex multi-dimensional instructions averaging 8 constraints per sample. (3) \textbf{Context Preparation}: We retrieve relevant documents using hybrid retrieval (BM25 + dense retrieval) and apply non-ideal simulation (Section 3.4) to create noise, knowledge gaps, and factual conflicts. (4) \textbf{Human Verification}: Instead of generating reference answers, the constructed samples undergo a rigorous two-stage expert verification process. Experts verify Instruction-Query Consistency and Context Validity (confirming topical relevance and the presence/absence of necessary information for non-ideal scenarios), filtering out low-quality samples to ensure ecological validity.

\subsection{Constraints Taxonomy}
\label{subsec:appendix A.3}

Table~\ref{tab:standard-subset-subdim} and figure~\ref{fig:subdimension_constraint_distribution} outlines the taxonomy across three dimensions. Persona Definition establishes the model's virtual identity and audience context, while Output Constraints dictate the structural format and stylistic boundaries. Crucially, the Knowledge Interaction Protocol enforces strict behavioral rules for evidence handling, such as citation and conflict resolution. This dimension moves beyond simple formatting to ensure the rigorous reliability required in enterprise environments.
\begin{table*}[t]
\centering
\caption{Detailed taxonomy and definitions of sub-dimension constraints. The table describes the specific requirements for Persona Definition, Output Constraints, and Knowledge Interaction Protocols used in the benchmark.}
\label{tab:standard-subset-subdim}
\small
\setlength{\tabcolsep}{5pt}
\renewcommand{\arraystretch}{1.15}
\begin{tabularx}{\textwidth}{@{} >{\centering\arraybackslash}p{1.8cm} >{\raggedright\arraybackslash}p{2.2cm} X @{}}
\toprule
\textbf{Category} & \textbf{Sub-Dimension} & \textbf{Description} \\
\midrule
\multirow{2}{*}{\makecell{Persona\\Definition}} 
  & Role 
  & Defines the user's virtual identity, such as a software engineer, medical expert, or customer service representative. \\
  & Audience 
  & Specifies the target readers of generated content, influencing the level of detail, terminology, and professionalism. \\
\midrule
\multirow{5}{*}{\makecell{Output\\Constraints}} 
  & Format 
  & Specifies the output structure, such as Markdown, JSON, or requirements like the number of sections to include. \\
  & Content 
  & Defines content requirements including word count limits, prefix/suffix specifications, and keyword frequency constraints. \\
  & Negative 
  & Explicitly specifies prohibited elements in the output, testing the model's fine-grained control capability. \\
  & Ordering 
  & Specifies the arrangement of output content, e.g., chronological order, alphabetical sorting, or frequency-based ranking. \\
  & Style 
  & Defines the response tone, ranging from rigorous and formal to relaxed and conversational. \\
\midrule
\multirow{5}{*}{\makecell{Knowledge\\Interaction\\Protocol}} 
  & Conflict Handling 
  & Defines how the model should respond when contradictory information exists within the knowledge base. \\
  & Knowledge Gap 
  & Requires the model to identify and explicitly report when necessary information is missing or unavailable. \\
  & Uncertainty Expr. 
  & Specifies how the model should express uncertainty when information is ambiguous or evidence is insufficient. \\
  & Source Filtering 
  & Instructs the model to selectively trust or ignore specific types of information sources. \\
  & Citation 
  & Requires key conclusions to be accompanied by directly cited original text excerpts as supporting evidence. \\
\bottomrule
\end{tabularx}
\end{table*}

\subsection{Domain Statistics}
\label{subsec:appendix A.4}

Table~\ref{tab:data_distribution} shows that the Medical and Financial domains exhibit the highest complexity, with 10.21 and 8.74 average constraints respectively. Legal tasks feature the highest density of Knowledge Protocol constraints (3.45). In contrast, Web Search peaks in Output Constraints (4.73) with lower protocol requirements (1.80), while Energy shows the lowest Persona usage (0.85).

\begin{figure}[!htb]
    \centering
  \includegraphics[width=\columnwidth]{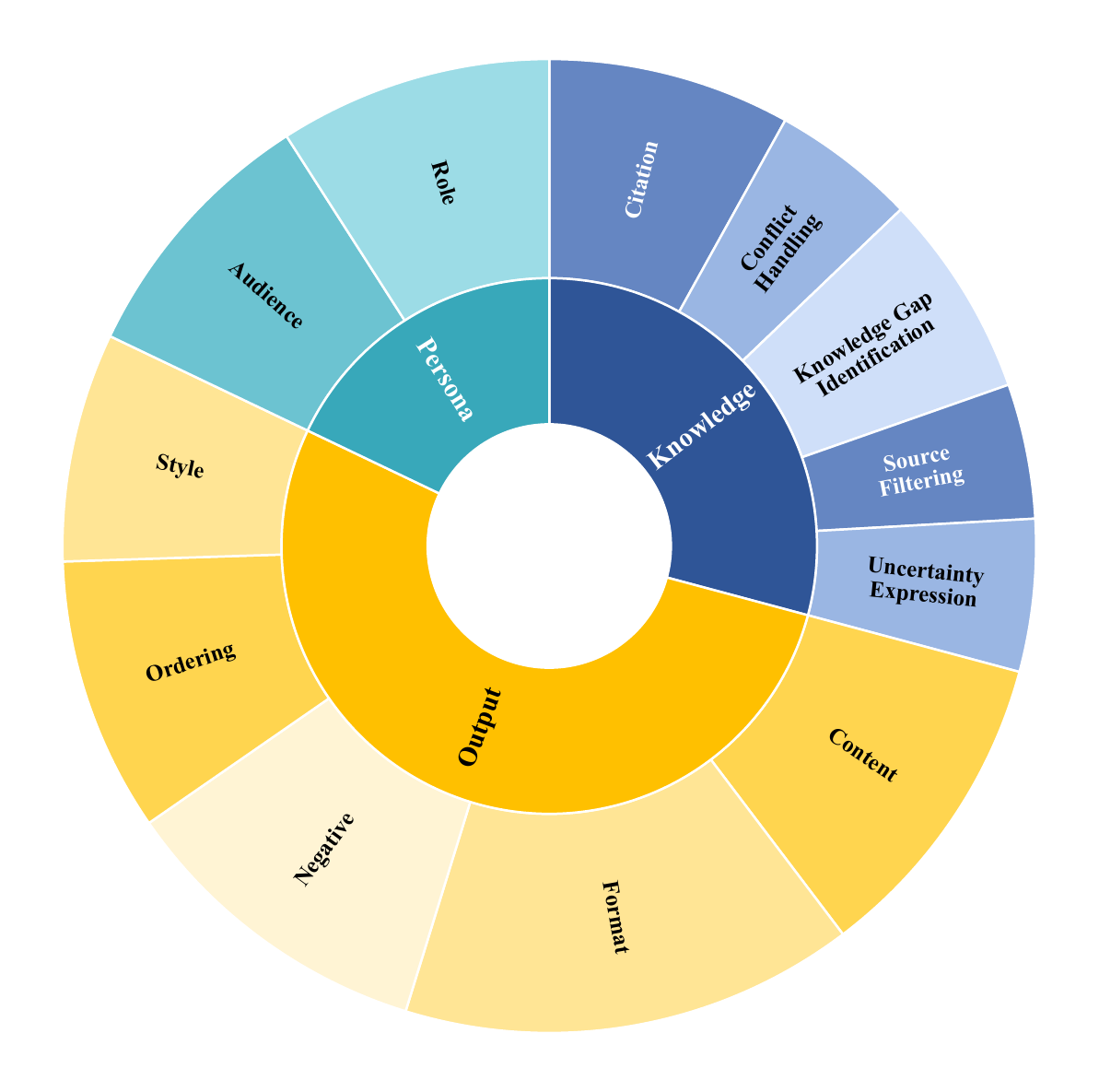}
  \caption{Hierarchical distribution of constraints in EnterpriseRAG. The inner ring represents the three primary categories, while the outer ring details the specific sub-dimensions.}
  \label{fig:subdimension_constraint_distribution}
\end{figure}

Figure~\ref{fig:subdimension_constraint_distribution}  illustrates the overall constraint distribution. Output Constraints constitute the majority (53.1\%), followed by Knowledge Interaction Protocols (29.2\%) and Persona Definitions. Within protocols, Citation and Conflict Handling represent the most frequent sub-dimensions.

\begin{table*}[t]
\centering
\caption{Statistics of queries and constraint density. The table shows the number of samples and the average number of constraints per query across the three orthogonal dimensions for each of the six vertical domains.}
\label{tab:data_distribution}
\small 
\setlength{\tabcolsep}{12pt} 

\begin{tabular}{lccccc}
\toprule
\textbf{Domain} &
  \textbf{\#Data} &
  \textbf{\#Constraints} &
  \textbf{\shortstack{Persona\\Def.}} &
  \textbf{\shortstack{Output\\Const.}} &
  \textbf{\shortstack{Knowledge\\Protocol}} \\
\midrule
\textbf{Financial}      & 100 & 8.74  & 1.94 & 5.08 & 1.72 \\
\textbf{Energy}         & 100 & 7.01  & 0.85 & 3.51 & 2.65 \\
\textbf{Legal}          & 44  & 8.61  & 1.59 & 3.57 & 3.45 \\
\textbf{Medical}        & 100 & 10.21 & 1.46 & 5.02 & 3.16 \\
\textbf{Party Building} & 48  & 7.79  & 1.21 & 4.08 & 2.50 \\
\textbf{Web Search}     & 99 & 8.41  & 1.87 & 4.73 & 1.80 \\
\midrule
\textbf{All}            & 491 & 8.40  & 1.50 & 4.46 & 2.45 \\
\bottomrule
\end{tabular}

\vspace{1mm}
\raggedright
\footnotesize
\textit{Note:} \#Data = Number of Data Samples; \#Const. = Avg. Constraints per query.
\end{table*}

\subsection{Non-Ideal Context Composition}
\label{subsec:appendix A.5}

Table~\ref{tab:Nonideal_distribution} reports the distribution of non-ideal context types in the final 983 instances. Knowledge gaps and factual conflicts are augmented due to sparsity in naturally occurring logs; we therefore explicitly report the natural vs. augmented breakdown for transparency.

\begin{table}[t]
\centering
\caption{Distribution of non-ideal contexts in EnterpriseRAG.}
\label{tab:Nonideal_distribution}

\resizebox{\columnwidth}{!}{%
\begin{tabular}{lcccc} 
\toprule
\textbf{Context} & \textbf{Count} & \textbf{Ratio} & \textbf{Natural Occ.} & \textbf{Augment.} \\
\midrule
Noisy             & 447 & 45.5\% & 447 (100\%) & -- \\
Knowledge Gaps    & 227 & 23.1\% & 12 (5.3\%) & 215 (94.7\%) \\
Factual Conflicts & 309 & 31.4\% & 27 (8.7\%)  & 282 (91.3\%) \\
\bottomrule
\end{tabular}%
}

\vspace{4pt} 
\parbox{\columnwidth}{
    \scriptsize 
    \textit{Note:} \textbf{Natural Occ.} = Natural Occurrence; \textbf{Augment.} = Augmentation. Percentages in parentheses indicate the proportion before augmentation. We augmented underrepresented failure modes to reflect realistic deployment distributions.
}
\end{table}


\section{Evaluation Reliability Analysis}
\label{sec:appendix B}
\subsection{Cross-Judge Consistency}
\label{subsec:appendix B.1}

We validate the stability of our evaluation protocol by comparing three LLM judges: Kimi-k2-thinking, Qwen3-235B-Thinking, and GPT-4o. As shown in Table~\ref{tab:evaluator_model_comparision}, the raw evaluation scores (upper section) exhibit minimal variance across judges; for instance, Loose IAS scores for DeepSeek-V3.1 differ by less than 0.5\% between Kimi and Qwen3.

This consistency is rigorously confirmed by statistical reliability metrics (lower section). We observe "Good" to "Excellent" reliability across all dimensions, with Intraclass Correlation Coefficients (ICC) exceeding 0.80 for Loose IAS and approaching 1.0 for robustness metrics. The high Spearman’s $\rho$ further indicates that different judges preserve the same relative model rankings, ensuring that our reported performance gaps are robust to the choice of evaluator.

\begin{table*}[t]
\centering
\caption{Consistency and Reliability Analysis of Evaluator LLMs. The upper section compares the raw evaluation scores of Kimi, Qwen3, and GPT-4o across three metrics. The lower section reports the statistical inter-annotator agreement metrics, validating the stability of the evaluation protocol. ICC values are reported with 95\% confidence intervals.}
\label{tab:evaluator_model_comparision}
\resizebox{0.95\textwidth}{!}{%
\begin{tabular}{lccccccccc}
\toprule
\multirow{2}{*}{\textbf{Model / Metric}} & \multicolumn{3}{c}{\textbf{Loose IAS}} & \multicolumn{3}{c}{\textbf{Reject Acc}} & \multicolumn{3}{c}{\textbf{Conflict Acc}} \\ \cmidrule(lr){2-4} \cmidrule(lr){5-7} \cmidrule(lr){8-10} 
 & Kimi & Qwen3 & GPT-4o & Kimi & Qwen3 & GPT-4o & Kimi & Qwen3 & GPT-4o \\ 
\midrule
\multicolumn{10}{c}{\textit{Raw Evaluation Scores (\%)}} \\ 
\midrule
Qwen3-235B-Thinking & 89.6 & 89.0 & 91.9 & 27.8 & 27.9 & 28.8 & 37.8 & 39.1 & 39.1 \\
DeepSeek-V3.1 & 86.0 & 86.2 & 90.4 & 18.9 & 19.4 & 19.4 & 28.3 & 28.9 & 31.0 \\
Gemini-2.5-Pro & 84.9 & 89.1 & 88.3 & 33.6 & 35.8 & 34.5 & 37.2 & 39.4 & 36.9 \\
GPT-4.1 & 82.7 & 83.8 & 85.8 & 10.1 & 9.3 & 9.3 & 16.6 & 16.6 & 17.6 \\ 
\midrule
\multicolumn{10}{c}{\textit{Inter-Judge Reliability Statistics}} \\ 
\midrule
\textbf{ICC (2,k)} & \multicolumn{3}{c}{0.809 \small{[0.11--0.99]}} & \multicolumn{3}{c}{0.999 \small{[0.99--1.00]}} & \multicolumn{3}{c}{0.996 \small{[0.98--1.00]}} \\
\textbf{Avg. Spearman's $\rho$} & \multicolumn{3}{c}{0.600} & \multicolumn{3}{c}{1.000} & \multicolumn{3}{c}{0.867} \\
\textbf{Reliability Verdict} & \multicolumn{3}{c}{\textbf{Good}} & \multicolumn{3}{c}{\textbf{Excellent}} & \multicolumn{3}{c}{\textbf{Excellent}} \\
\bottomrule
\end{tabular}%
}
\end{table*}

\subsection{Synthetic vs. Real-world Data Performance}
\label{subsec:appendix B.2}

\begin{figure*}[t]
    \centering
    \includegraphics[width=0.85\textwidth]{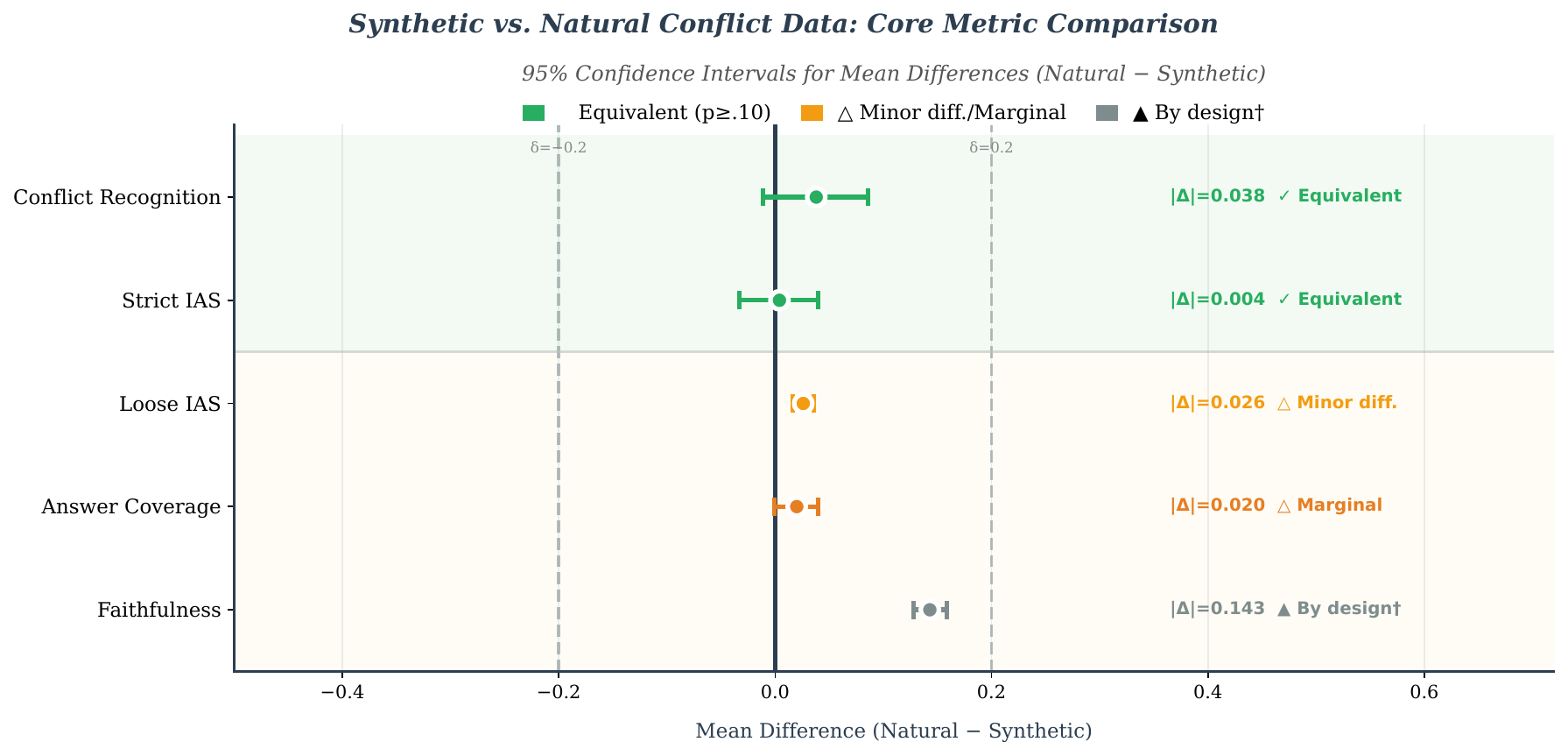}
    \caption{Equivalence testing results comparing synthetic and natural conflict data across five evaluation metrics. Points represent mean differences (Natural - Synthetic) with 95\% confidence intervals; dashed lines mark equivalence bounds (±0.2). Core robustness metrics (Conflict Recognition and Strict IAS) achieve statistical equivalence, validating the synthetic data generation approach.}
    \label{fig:conflict_natural_vs_syn}
\end{figure*}

 Figure~\ref{fig:conflict_natural_vs_syn} and Table~\ref{tab:synthetic-validation-fix} contrast model performance on naturally occurring (n=27) versus synthetic (n=282) conflicts. Across five models, synthetic samples show slightly lower Faithfulness ($\Delta$ = -0.143) and Answer Coverage ($\Delta$ = -0.020) due to their engineered contradictions. Critically, we observe statistical equivalence in the core robustness metrics of Conflict Recognition Accuracy and Strict IAS ($\Delta$ = -0.038, p = .120; $\Delta$ = -0.004, p = .826), with 95\% confidence intervals within ±0.2. This confirms that our synthesis pipeline effectively replicates the difficulty of real-world scenarios, ensuring the ecological validity of the augmentation strategy.

\begin{table}[t] 
\centering
\caption{Natural vs. Synthetic Data Comparison. \textbf{n.s.} denotes no significant difference ($p \ge 0.05$), indicating successful replication of difficulty on core metrics.}
\label{tab:synthetic-validation-fix}
\resizebox{\columnwidth}{!}{%
\begin{tabular}{lcccc}
\toprule
\textbf{Metric} & \textbf{Natural} & \textbf{Synth.} & \textbf{Diff.} & \textbf{\textit{p}-val} \\
\midrule
\multicolumn{5}{l}{\textit{\textbf{Core Robustness} (Target: Equivalent)}} \\
Conflict Recog. & 0.360 & 0.322 & -0.038 & .120 \textbf{(n.s.)} \\
Strict IAS      & 0.160 & 0.156 & -0.004 & .826 \textbf{(n.s.)} \\
\midrule
\multicolumn{5}{l}{\textit{General Quality}} \\
Loose IAS       & 0.793 & 0.767 & -0.026 & <.001$^{***}$ \\
Answer Cov.     & 0.551 & 0.532 & -0.020 & .054 \phantom{$^{***}$} \\
Faithfulness    & 0.827 & 0.684 & -0.143 & <.001$^{***}$ \\
\bottomrule
\end{tabular}%
}
\end{table}

\subsection{Human–LLM Judge Alignment Study}
\label{subsec:appendix B.3}

To ensure rigorous evaluation, we employed a two-stage annotation protocol on a stratified sample of 150 instances. First, two experts independently labeled the data, achieving robust Inter-Annotator Agreement (IAA)  (avg. $\kappa$=0.85, see Table~\ref{tab:human_llm_alignment}), which validates the clarity of our instruction taxonomy. Disagreements were adjudicated to establish a Gold Standard.

Against this baseline, the LLM judge demonstrates substantial reliability with an average $\kappa$ of 0.77. Conflict Recognition achieves the highest alignment ($\kappa$=0.93) due to the objective nature of contradiction detection, while Strict Adherence  ($\kappa$=0.68) exhibits minor divergence on borderline formatting nuances.

\begin{table}[t]
\centering
\small 
\setlength{\tabcolsep}{6pt} 
\caption{\textbf{Human-LLM Judge Alignment Study.} Analysis based on a stratified sample of 150 instances (50 per dimension). \textbf{Data Quality}: Inter-Annotator Agreement (IAA) between two human experts. \textbf{LLM Judge Reliability}: Primary evaluator (Kimi-k2-thinking) vs. adjudicated gold standard.}
\label{tab:human_llm_alignment}
\resizebox{\columnwidth}{!}{%
\begin{tabular}{lccc}
\toprule
\multirow{2}{*}{\textbf{Evaluation Dimension}} & \textbf{Data Quality} & \multicolumn{2}{c}{\textbf{LLM Judge Reliability}} \\ 
\cmidrule(lr){2-2} \cmidrule(lr){3-4} 
 & Human IAA & Cohen's $\kappa$ & Agreement \\ 
 & ($\kappa$) &  & Rate (\%) \\ 
\midrule
Strict IAS & 0.73 & 0.68 & 86 \\ 
Proper Rej. & 0.87 & 0.71 & 87 \\ 
Conflict Recog. & 0.96 & 0.93 & 91 \\ 
\midrule
\textbf{Overall avg.} & \textbf{0.85} & \textbf{0.77} & \textbf{88} \\ 
\bottomrule
\end{tabular}%
}
\end{table}

\section{Statistical Analysis Details}
\label{sec:appendixStats}

\subsection{Figure~\ref{fig:conflict_coverage_corr}}

\paragraph{Data and Method}
Each point: model-level aggregate over $n{=}309$ conflict instances (X: conflict recognition rate; Y: mean answer coverage). We compute \textbf{Spearman's $\rho$} separately for reasoning-enhanced ($n{=}8$) and standard ($n{=}5$) models due to their distinct architectures.

\paragraph{Limitations}
Small sample sizes limit power. Models within families (e.g., Qwen3) may not be fully independent; sensitivity analysis yields consistent patterns. Model-level correlations may not reflect instance-level relationships.

\subsection{Figure~\ref{fig:reason_vs_instruct}} Method: McNemar's test (one-sided) for paired binary outcomes. Each instance evaluated by both reasoning-enhanced and standard models within the same family.

Samples: Noisy Retrieval ($n{=}447$), Knowledge Gaps ($n{=}227$), Factual Conflicts ($n{=}309$). Exact binomial test when discordant pairs $< 25$; otherwise z-test with continuity correction.

Confidence Intervals: Percentile bootstrap (10,000 resamples) preserving pairing.

Multiple Testing: Uncorrected $p$-values for 10 planned comparisons; Bonferroni correction ($\alpha{=}.005$) does not change conclusions.

\subsection{Figure ~\ref{fig:robustness_woconstrait_compare}} We compare accuracy between independent groups (samples with vs. without the protocol) using standard methods for comparing two proportions.

For each model, we construct a $2{\times}2$ contingency table and apply: \textbf{Pearson's $\chi^2$ test} (with Yates' correction) when expected counts $\geq 5$, \textbf{Fisher's exact test} otherwise.

We report the accuracy difference $\Delta = \hat{p}_{\text{with}} - \hat{p}_{\text{without}}$ with 95\% Wald confidence intervals.

Sample sizes: rejection (157 with / 69 without), conflict recognition (113 / 193). Analysis used SciPy. Of 26 tests, 25 used $\chi^2$ and 1 used Fisher's exact. Uncorrected $p$-values are reported; Bonferroni correction ($\alpha = .002$) does not alter substantive conclusions.

\subsection{Figure~\ref{fig:conflict_natural_vs_syn}}

We use \textbf{TOST equivalence testing} to validate that synthetic conflicts ($n{=}282$) replicate natural conflict difficulty ($n{=}27$). Equivalence margin: $\delta{=}{\pm}0.2$ (20pp, based on RAG benchmark reliability thresholds). Decision rule: 95\% CI for difference (Natural $-$ Synthetic) must fall within $[-0.2, 0.2]$.

Aggregate mean accuracy across 5 models for each dataset; bootstrap 95\% CI (10,000 resamples).

\paragraph{Limitations}
Small natural sample ($n{=}27$); model-level aggregation; families share architectures.

\subsection {Table~\ref{tab:synthetic-validation-fix}}

We use \textbf{paired-samples $t$-test} to compare 13 models' performance on natural ($n=27$) vs. synthetic ($n=282$) conflicts, with Cohen's $d$ for effect sizes.

\paragraph{Limitations}
Models within families may not be fully independent; unequal natural/synthetic sample sizes affect precision but not paired comparison validity.

\section{Additional Experiments}
\label{sec:appendix C}

\subsection{Fine-grained Analysis}
\label{subsec:appendix C.1}

\begin{figure*}[!htb]
    \centering
    \includegraphics[width=0.7\textwidth]{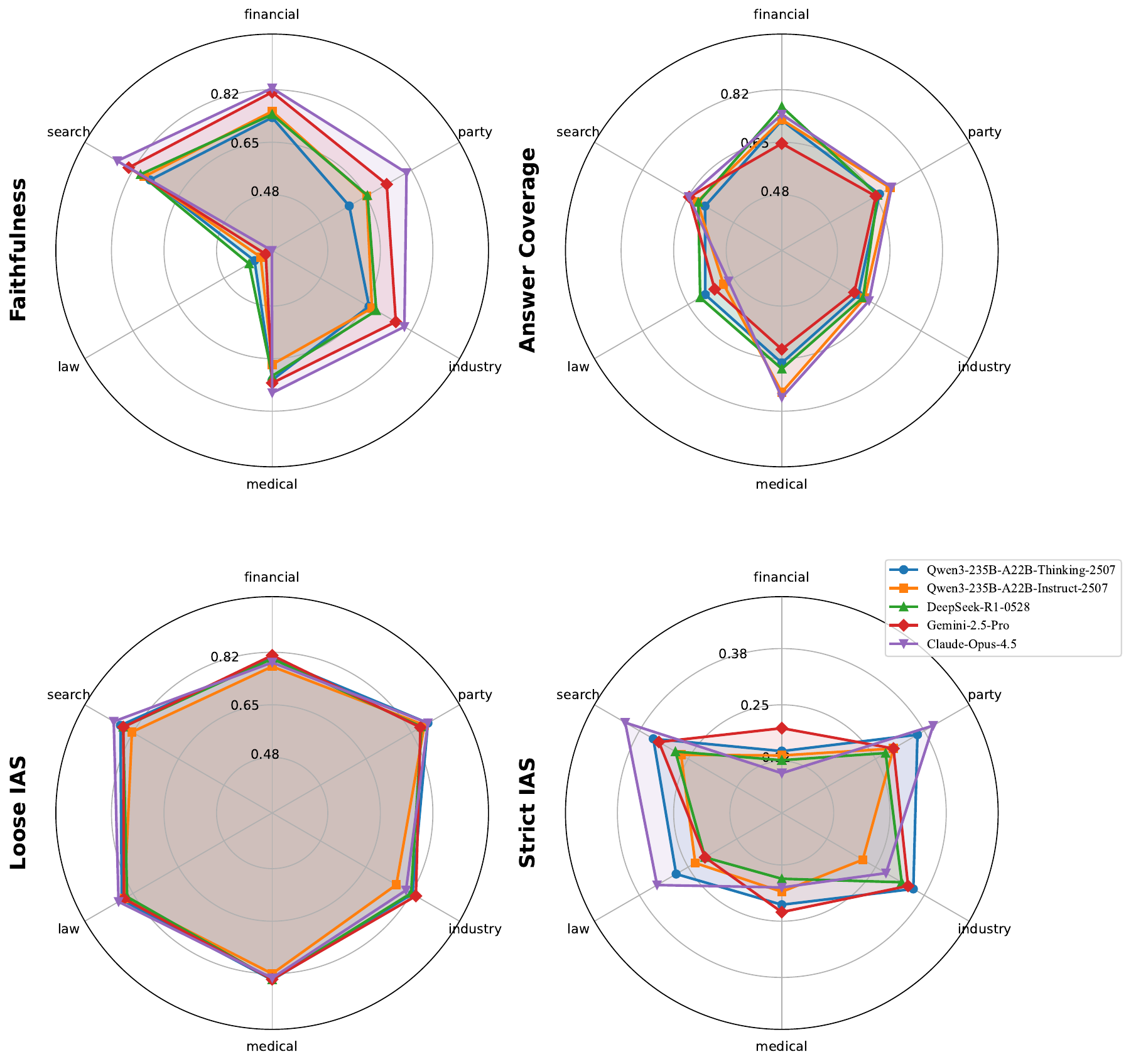}
    \caption{Model performance on noisy retrieval contexts across six vertical domains. Radar charts illustrate the trade-offs between Faithfulness, Answer Coverage, and Instruction Adherence (Strict/Loose) for representative models.}
    \label{fig:domain_performance_radar}
\end{figure*}

Figure~\ref{fig:domain_performance_radar} illustrates the performance of five representative LLMs—including both reasoning-enhanced and standard instruction-tuned variants—across the six vertical domains of EnterpriseRAG under noisy retrieval conditions. Across all domains, models generally maintain high Faithfulness and Loose IAS , but experience a sharp "orchestration collapse" in Strict IAS, confirming that simultaneously satisfying multiple domain-specific constraints remains a primary bottleneck. Specifically, domains with higher complexity and stricter behavioral requirements, such as Medical and Legal, exhibit lower absolute Strict IAS scores compared to Web Search. This trend aligns with the domain statistics in Table~\ref{tab:data_distribution}, which show that Medical and Legal tasks feature the highest average number of constraints and the densest concentration of Knowledge Interaction Protocols. Furthermore, reasoning-enhanced models (e.g., Qwen3-235B-Thinking and DeepSeek-R1) consistently outperform their counterparts across all metrics and domains, particularly in Strict IAS and Answer Coverage. These results suggest that the difficulty of instruction adherence is inherently tied to domain-specific constraint density, and that robust performance in complex enterprise scenarios is an emergent capability heavily dependent on inference-time reasoning rather than simple pattern matching.

\section{Complete Prompts Repository}
\label{sec:appendix D}

Note: All prompts and examples presented in this paper have been translated from the original Chinese for readership clarity. The actual evaluation was performed using the Chinese versions.

\subsection{The Prompts for Data Construction}
\label{subsec:appendix D.1}

\begin{figure*}[t] 
    \centering
    \raggedright \large \textbf{D.1.1 \quad Initial Atomic Libraries Generation} \vspace{0.5em}
    
    \begin{tcolorbox}[
        colback=white,
        colframe=black!80,
        boxrule=0.8pt,
        arc=2mm,
        left=8pt, right=8pt, top=8pt, bottom=8pt, 
        fontupper=\footnotesize 
    ]
        \textbf{[System]} \\
        You are an expert in constructing the EnterpriseRAG benchmark. Your goal is to generate realistic, complex atomic instructions for retrieval-augmented generation systems in specific domains.
        
        \vspace{0.5em}
        
        \textbf{[Context Input]} \\
        \textbf{Definition of TaskRAG Dimensions:} 
        \begin{itemize}[leftmargin=1em, labelsep=0.5em, nosep] 
            \item \textbf{P (Persona):} Role, Audience.
            \item \textbf{C (Constraints):} Format, Content, Negative (forbidden), Ordering, Style.
            \item \textbf{K (Knowledge Protocol):} Conflict Handling, Gap ID, Uncertainty, Source Filtering, Evidence Chaining.
        \end{itemize}
        
        \vspace{0.5em}

        \textbf{[Task Description]} \\
        Domain: \bluevar{scenario}. Details: \bluevar{detailed\_scenario\_description}. \\
        \textbf{Requirements:}
        \begin{enumerate}[leftmargin=1.5em, itemsep=0pt, topsep=0pt, parsep=0pt] 
            \item Generate 5+ atomic instructions per sub-dimension (except Operational Goal).
            \item Assign probability (0.0-1.0) and verify \textbf{Generalization} (vs context-specific).
            \item Determine \textbf{Evaluability} (Rule vs LLM) and provide \textbf{Judge Logic} (ifeval/prompt).
            \item Refer to \textbf{Instruct\_Follow\_Evaluation} for constraint types (word count, bullets, etc.).
        \end{enumerate}

        \vspace{0.5em}

        \textbf{[Response Examples]} \\
        Output strictly in the following JSON format:

\begin{minipage}{\linewidth}
\begin{lstlisting}[language=json]
{
  "Persona & Scenario": {
    "Role": [
      {
        "instruction": "Do not include names of medical staff other than the attending physician.",
        "probability": 0.4,
        "generalization": "no",
        "judge_method": "LLM",
        "judge_detail": "Prompt: Determine if report contains other names... Format: {judge_format}..."
      },
      {
        "instruction": "Do not use exclamation marks.",
        "probability": 0.5,
        "generalization": "yes",
        "judge_method": "rule",
        "judge_detail": { "instruction_id_list": ["forbidden_words"], "kwargs": [{ "forbidden_words": ["!"] }] }
      }
    ]
  }
}
\end{lstlisting}
\end{minipage}

    \end{tcolorbox}
    \caption{The prompt template used for generating EnterpriseRAG Atomic Libraries.}
    \label{fig:Initial_Atomic_Libraries_Generation}
\end{figure*}

\begin{figure*}[t] 
    \centering
    \raggedright \large \textbf{D.1.2 \quad Check Combined Constraints Conflict} \vspace{0.5em}
    
    \begin{tcolorbox}[
        colback=white,
        colframe=black!80,
        boxrule=0.8pt,
        arc=2mm,
        left=8pt, right=8pt, top=8pt, bottom=8pt, 
        fontupper=\footnotesize 
    ]
        Please evaluate whether there are logical conflicts or contradictions within the atomic instruction list below. If conflicts exist, prioritize retaining content related to \textbf{[Operational Execution]}.

        \vspace{0.8em}
        
        \textbf{Atomic Instruction List:}
        \bluevar{constraints\_list}
        
        \vspace{0.5em}

        Please strictly return the following JSON:

\begin{lstlisting}[language=json]
{
  "has_conflict": false,
  "reason": "Brief explanation",
  "unconflict_list": "If conflicts exist and can be resolved by removing specific atomic instructions, provide the list of indices after removal, e.g., [1, 3, 5]; otherwise, provide the original list of atomic instruction indices."
}
\end{lstlisting}

    \end{tcolorbox}
    \caption{The prompt template used for detecting conflict in combined constraints.}
    \label{fig:Check_Combined_Constraits_Conflict}
\end{figure*}

\begin{figure*}[t] 
    \centering
    \raggedright \textbf{D.1.3 \quad Prompt Templates for Instruction Refinement} \vspace{0.3em}
    
    \begin{tcolorbox}[
        colback=white, colframe=black!80, boxrule=0.8pt, arc=2mm,
        left=8pt, right=8pt, top=8pt, bottom=8pt,
        fonttitle=\bfseries\small, title=Prompt 1: Filter Irrelevant Instructions
    ]
\footnotesize
        Based on the given question, please remove atomic instructions that cannot be triggered by the current data:

        \vspace{0.3em}
        \textbf{Original Atomic Instruction List:} \\
        \bluevar{constraints\_list}

        \textbf{Question:} \pyvar{question} \\
        \textbf{Context:} \pyvar{context}

        Please analyze which atomic instructions cannot be triggered in the current question/context and remove them.
        Please return in the following JSON format:

\begin{lstlisting}[language=json]
{
  "removed_atoms": ["List of indices of removed atomic instructions, e.g., [1, 3]"],
  "reason": "Reason for removal"
}
\end{lstlisting}
        If no atomic instructions need to be removed, please return an empty list.
    \end{tcolorbox}

    \vspace{0.5em}

    \begin{tcolorbox}[
        colback=white, colframe=black!80, boxrule=0.8pt, arc=2mm,
        left=8pt, right=8pt, top=8pt, bottom=8pt,
        fonttitle=\bfseries\small, title=Augment Constraints Based on the Query]
        \footnotesize
        Based on the given question and context, please add specific constraints to the current instruction list:

        \textbf{Current Atomic Instruction List:} \\
        \bluevar{constraints\_list}

        \textbf{Question:} \pyvar{question} \quad \textbf{Context:} \pyvar{context}

        \textbf{Requirements are as follows:}
        \begin{enumerate}[leftmargin=1.2em, itemsep=0pt, topsep=2pt, parsep=0pt]
            \item Analyze the question and context features; add 1-2 new constraints to existing atomic instructions.
            \item Must not repeat or overlap with existing atomic instructions.
            \item Consider the following dimensions for new constraints:
            \begin{itemize}[leftmargin=1em, nosep]
                \item \textbf{Content \& Format:} Format/Content/Role/Style/Tone/Length, etc.
                \item \textbf{Positive \& Negative:} Must include/Must not include/Avoid including, etc.
                \item \textbf{Knowledge Interaction Protocol:} Citation/No Citation/Prioritize Citation/Conflict Handling/Missing Info Handling/Uncertainty Expression.
            \end{itemize}
            \item New constraints must be specific and actionable (evaluable via code or LLM) and avoid overly broad or vague descriptions that make evaluation difficult.
            \item New constraints should be relevant to the current question/context and improve answer quality, but description should not be too detailed (specific only to current context); it should have some generalization.
            \item If the current atomic instruction list is sufficiently complete, no constraints need to be added.
            \item Please return in the following JSON format:
        \end{enumerate}

\begin{lstlisting}[language=json]
{
  "additional_constraints": [
     {
      "category": "Additional Constraints",
      "dimension": "Additional Constraints",
      "instruction": "Constraint instruction text",
      "judge_method": "llm",
      "judge_detail": "LLM-as-judge prompt for evaluating this constraint"
     }
  ],
  "reason": "Reason for adding these constraints"
}
\end{lstlisting}
        If no constraints need to be added, please return an empty list.

        \textbf{8. Reference Example:}
\begin{lstlisting}[language=json]
{
  "instruction": "Do not include names of medical staff other than the attending physician in the report.",
  "probability": 0.4, "generalization": "no", "judge_method": "LLM",
  "judge_detail": "Please determine whether the report below only contains the attending physician's name... Output format:\n{judge_format}\n\nReport:\n{response}\n\nOriginal medical record:\n{context}"
}
\end{lstlisting}
        
        \textbf{9.} \texttt{judge\_detail} must be complete and accurate. It must explicitly specify the evaluation model's output as \pyvar{judge\_format}, where \texttt{judge\_format} is \texttt{\{\{"Does it satisfy instruction constraints": "Yes or No", "Reason": "Provide judgment reason"\}\}}. The evaluation can cite \texttt{context}, \texttt{question}, and \texttt{response} fields. The instruction and \texttt{judge\_detail} must maintain logical consistency.

    \end{tcolorbox}
    \label{fig:query_refinement_prompts}
\end{figure*}

\begin{figure*}[t] 
    \centering
    \raggedright \textbf{D.1.4 \quad Quality Assessment of Complex Instructions} \vspace{0.3em}
    
    \begin{tcolorbox}[
        colback=white,
        colframe=black!80,
        boxrule=0.8pt,
        arc=2mm,
        left=8pt, right=8pt, top=8pt, bottom=8pt,
        fontupper=\small 
    ]
        You are an expert in benchmarking \bluevar{task\_name} tasks. Your task is to evaluate the quality of the following complex instruction, composed of multiple randomly combined atomic instructions, to determine if it is suitable as a valid evaluation case.

        \vspace{0.5em}
        
        \textbf{Background:} \\
        This instruction is used to evaluate a \bluevar{task\_name} large model oriented towards the \pyvar{domain} domain. The model needs to answer questions based on the given context while strictly adhering to the instructions.

        \vspace{0.5em}
        
        \textbf{Atomic Instructions to be Evaluated:} \\
        \bluevar{instruction\_text}

        \vspace{0.5em}
        
        \textbf{Please evaluate the above combined instruction based on the following five dimensions and output your analysis results in JSON format:}

        \begin{enumerate}[leftmargin=1.2em, itemsep=2pt, topsep=2pt, label=\textbf{\arabic*.}]
            \item \textbf{Logical Consistency}: 
            Evaluate whether there are internal conflicts or incoordination among the parts of the instruction (role, goal, format, constraints). Score (1-5, where 1 is severely inconsistent, 5 is completely consistent).
            
            \item \textbf{Feasibility}: 
            Evaluate whether it is theoretically possible for a top-tier \bluevar{task\_name} model to satisfy all these requirements simultaneously. Check for absolute contradictions (e.g., requiring a list while prohibiting lists). Score (1-5, where 1 is completely unexecutable, 5 is completely executable).
            
            \item \textbf{Realism}: 
            Evaluate whether this instruction combination simulates a real, reasonable work scenario likely to occur in a \bluevar{task\_name} task within the \bluevar{domain} domain. Score (1-5, where 1 is completely unrealistic, 5 is very realistic).
            
            \item \textbf{Clarity of Evaluation}: 
            Evaluate whether we still have clear, actionable methods (whether via rules or LLM-as-Judge) to judge if the model followed every instruction after combination. Score (1-5, where 1 is very vague evaluation criteria, 5 is very clear).
            
            \item \textbf{Appropriate Complexity}: 
            Evaluate whether the overall difficulty is too simple, moderate, or too complex to be practical. Score (1-5, where 1 is too simple/complex making it ineffective, 5 is moderate complexity with good discrimination).
        \end{enumerate}

        \vspace{0.5em}

        \textbf{Output Format:} \\
        Please strictly return your evaluation results following the JSON structure below.

\begin{lstlisting}[language=json]
{
  "instruction_id": "[Assign a unique ID for this instruction combination]",
  "evaluation_summary": {
    "logical_consistency": {
      "score": <Score int>,
      "reasoning": "<Your analysis reasoning>"
    },
    "feasibility": {
      "score": <Score int>,
      "reasoning": "<Your analysis reasoning; explicitly point out contradictions if any>"
    },
    "realism": {
      "score": <Score int>,
      "reasoning": "<Your analysis reasoning>"
    },
    "evaluation_clarity": {
      "score": <Score int>,
      "reasoning": "<Your analysis reasoning>"
    },
    "complexity": {
      "score": <Score int>,
      "reasoning": "<Your analysis reasoning>"
    }
  },
  "overall_judgment": {
    "average_score": <Composite Score float>,
    "recommendation": "<'Recommended' | 'Use with Caution' | 'Not Recommended'>",
    "final_remarks": "<Final summary and modification suggestions for this instruction>"
  }
}
\end{lstlisting}

    \end{tcolorbox}
    \label{fig:quality_assessment_prompt}
    \caption{The prompt template for evaluating and scoring the refined complex instruction.}
\end{figure*}

\begin{figure*}[t] 
    \centering
    \raggedright \textbf{D.1.5 \quad Document Relevance Assessment} \vspace{0.3em}
    
    \begin{tcolorbox}[
        colback=white,
        colframe=black!80,
        boxrule=0.8pt,
        arc=2mm,
        left=8pt, right=8pt, top=8pt, bottom=8pt,
        fontupper=\small 
    ]
        \textbf{[System Instruction]} \\
        \textbf{\# Task Description} \\
        Please strictly evaluate whether the retrieved document below effectively supports answering the user's question. Analyze the relevance between the document and the question step-by-step and output structured results.

        \vspace{0.3em}
        
        \textbf{\# Evaluation Steps}
        \begin{enumerate}[leftmargin=1.2em, itemsep=0pt, topsep=2pt, parsep=0pt]
            \item \textbf{Understand Question Core}
            \begin{itemize}[leftmargin=1em, nosep]
                \item Extract keywords and core requirements of the user question, clarifying the type of information needed for the answer (e.g., data, reasons, steps, etc.).
            \end{itemize}
            
            \item \textbf{Document Content Analysis}
            \begin{itemize}[leftmargin=1em, nosep]
                \item Check the document sentence by sentence, marking content directly related to the question (e.g., data, definitions, causal explanations).
                \item Identify potentially indirectly supporting information (e.g., background knowledge, analogous cases).
            \end{itemize}
            
            \item \textbf{Relevance Judgment}
            \begin{itemize}[leftmargin=1em, nosep]
                \item Determine if the document contains \textbf{key evidence} needed to answer the question (e.g., "Yes/No", must specify concretely).
                \item Check if the information is complete and reliable (e.g., source of data, existence of contradictions).
            \end{itemize}
            
            \item \textbf{Support Confirmation}
            \begin{itemize}[leftmargin=1em, nosep]
                \item Confirm whether the document content can fully or partially answer the question, rather than being unable to answer or completely irrelevant to the question content.
            \end{itemize}
        \end{enumerate}

        \vspace{0.3em}

        \textbf{\# Output Format}
\begin{lstlisting}[language=json]
{
  "supports_question": "Yes/No",
  "confidence": "Percentage (0-100%)",
  "reasoning": "1-2 sentences explaining the basis, e.g., missing information or completely irrelevant to question content",
  "key_evidence_citations": ["Original text fragments from the document"]
}
\end{lstlisting}

        \vspace{0.8em}
        \hrule 
        \vspace{0.8em}

        \textbf{[User Instruction]} \\
        \textbf{**Data for Analysis**} \\
        User Question: \pyvar{0} \\
        Retrieved Document: \pyvar{1}

    \end{tcolorbox}
    \caption{The prompt used to verify the relevance of retrieved documents and query.}
    \label{fig:relevance_verify_prompt}
\end{figure*}

\begin{figure*}[t] 
    \centering
    \raggedright \textbf{D.1.6 \quad Conflict Management} \vspace{0.3em}
    
    \begin{tcolorbox}[
        colback=white, colframe=black!80, boxrule=0.8pt, arc=2mm,
        left=8pt, right=8pt, top=8pt, bottom=8pt,
        fonttitle=\bfseries\small, title=Conflict Detection
    ]
        \footnotesize
        \textbf{[System]} \\
        You are an expert in information consistency and time-sensitivity analysis. Please judge whether conflicts/contradictions exist in the given context segments: two or more segments provide opposite or mutually exclusive conclusions regarding the same fact.
        Please list:
        1) Whether the above issue exists (Yes/No);
        2) The specific context segments involved in the conflict, labeled as Index 1 and Index 2 (indices start from 0);
        3) Summary of the conflict point;
        4) Whether it affects the answer (Yes/No) and the reason.

        \vspace{0.5em}
        \textbf{[User]} \\
        User Question: \pyvar{query} \\
        Context: \\
        \texttt{[0] Context\_0} \\
        \texttt{[1] Context\_1} ...

        Please output in JSON:
        \texttt{\{"has\_conflict": "Yes/No", "details": [\{"index1":[], "index2":[], "summary":"...", "affects\_answer": "Yes/No"\}]\}}
    \end{tcolorbox}

    \vspace{0.5em}

    \begin{tcolorbox}[
        colback=white, colframe=black!80, boxrule=0.8pt, arc=2mm,
        left=6pt, right=6pt, top=6pt, bottom=6pt,
        fonttitle=\bfseries\small, title=Conflict document synthesis]
        \footnotesize
        \textbf{[System]} You are a data construction expert.

        \vspace{0.3em}
        \textbf{[User]} 
        Please carefully read and follow the instructions below to complete a conflict context construction task.

        \textbf{\# Task Objective} \\
        Your task is to act as a data fabrication expert. Based on the user's "Question" and a series of "Correct Context" segments, you need to construct \pyvar{num} new, deceptive context segments. These new segments must contain information that directly conflicts or contradicts specific facts in the "Correct Context".

        \textbf{\# Core Requirements}
        \begin{enumerate}[leftmargin=1.2em, itemsep=0pt, topsep=2pt, parsep=0pt]
            \item \textbf{Modify Based on Facts}: Do not fabricate information out of thin air that is irrelevant to the original context. You must select one or more key fact points (e.g., numbers, dates, names, conclusions, status) from the "Correct Context" and modify them to create contradictions.
            \item \textbf{Maintain Context Relevance}: The constructed conflict segments must be highly relevant to the original question and context in terms of topic and phrasing. They should read naturally and credibly, not appearing obviously fake.
            \item \textbf{Explicitly Identify Conflict Points}: After construction, clearly indicate which original segment(s) the new segment conflicts with and concisely summarize the core content of the conflict.
        \end{enumerate}

        \textbf{\# Input Format}
        \begin{itemize}[leftmargin=1em, nosep]
            \item \textbf{Question}: User's original question.
            \item \textbf{Correct Context}: One or more segments, each prefixed with \texttt{-[Index]}, e.g., \texttt{-[0]}.
        \end{itemize}

        \textbf{\# Output Format} \\
        Strictly output in the following JSON format without additional explanation:
\begin{lstlisting}[language=json]
{
  "generated_contexts": [
    {
      "source_indices": [0],
      "conflict_id": "c0",
      "text": "Your first constructed conflict segment here. It should look credible but contain information contradicting paragraph `-[0]`.",
      "conflict_summary": "E.g.: Changed the release year 2023 in the original context to 2022."
    },
    {
      "source_indices": [1, 2],
      "conflict_id": "c1",
      "text": "Your second constructed conflict segment...",
      "conflict_summary": "E.g.: Replaced the main contributor 'John Doe' with 'Jane Doe'."
    }
  ]
}
\end{lstlisting}
        
        \textbf{Usage Example:} \\
        \textbf{Question:} \pyvar{query} \quad \textbf{Correct Context:} \pyvar{context} \quad \textbf{Output:}
    \end{tcolorbox}
    \caption{The two-stage process for conflict management: detection of existing contradictions and generation of synthetic conflicts to test model robustness.}
    \label{fig:conflict_management}
\end{figure*}

\subsection{The Prompts for Evaluation}
\label{subsec:appendix D.2}

\begin{figure*}[t] 
    \centering
    \raggedright \textbf{D.2.1 \quad Answer Coverage Evaluation} \vspace{0.3em}
    
    \begin{tcolorbox}[
        colback=white, colframe=black!80, boxrule=0.8pt, arc=2mm,
        left=8pt, right=8pt, top=8pt, bottom=8pt,
        fonttitle=\bfseries\small, title=Prompt 1: Claim Extraction
    ]
        \footnotesize
        You are a highly precise information extraction expert. Your task is to analyze a user's question and a provided context, then extract all relevant factual claims. You must classify each claim as either \texttt{[core]} or \texttt{[supplementary]}.

        \begin{itemize}[leftmargin=1em, nosep]
            \item \texttt{[core]}: The essential, direct answer to the question.
            \item \texttt{[supplementary]}: Valuable, additional information like preconditions, exceptions, timelines, or problem-solving steps.
        \end{itemize}

        Follow the output format exactly as shown in the examples. Each claim must be on a new line. If no relevant information exists, you MUST respond with the single word: "None".

        \vspace{0.3em}
        \hrule
        \vspace{0.3em}

        \textbf{--- Example 1 ---} \\
        \textbf{Question:} How do I reset my password? \\
        \textbf{Context:} To reset your password, click the 'Forgot Password' link on the login page. You will receive an email with instructions. Please note that the reset link is only valid for 10 minutes. \\
        \textbf{Key Information Points:} \\
        \texttt{[core]} Users can reset their password by clicking the 'Forgot Password' link. \\
        \texttt{[supplementary]} After clicking the link, an email with instructions will be sent. \\
        \texttt{[supplementary]} The password reset link is valid for only 10 minutes.

        \vspace{0.3em}
        \textbf{--- Example 2 ---} \\
        \textbf{Question:} What happens if the 'Pay without getting out' button doesn't respond in the app? \\
        \textbf{Context:} In the 'Pay without getting out' section of the Alipay mini-program, you need to swipe up to reveal the fuel pump selection screen. If that doesn't work, ensure your network connection is stable. For persistent issues, contact support at 400-123-4567. \\
        \textbf{Key Information Points:} \\
        \texttt{[core]} The user needs to swipe up on the screen to show the fuel pump selection page. \\
        \texttt{[supplementary]} The user should check if their network connection is stable. \\
        \texttt{[supplementary]} For persistent issues, users can contact support at 400-123-4567. \\
        \textbf{--- End of Examples ---}

        \vspace{0.5em}
        Now, perform the task for the following real data, note that there may be more than one core claim involved.

        \textbf{Question:} \pyvar{query} \\
        \textbf{Context:} \pyvar{context} \\
        \textbf{Key Information Points:}
    \end{tcolorbox}

    \vspace{0.5em}

    \begin{tcolorbox}[
        colback=white, colframe=black!80, boxrule=0.8pt, arc=2mm,
        left=8pt, right=8pt, top=8pt, bottom=8pt,
        fonttitle=\bfseries\small, title=Prompt 2: Coverage Verification
    ]
        \footnotesize
        You are a meticulous verifier. For the given "Generated Answer", determine if each "Key Information Statement" from the list is semantically covered.

        \textbf{Generated Answer:} \\
        \pyvar{answer}

        \textbf{Key Information Statements:} \\
        \pyvar{claims}

        Respond ONLY with a JSON object where keys are the exact claim index number(start from 1) and values are a boolean (true for covered, false for not covered).
        Example: \texttt{\{"1": true, "2": false\}}

        \textbf{JSON Output:}
    \end{tcolorbox}
    \label{fig:coverage_eval_prompts}
\end{figure*}

\begin{figure*}[t] 
    \centering
    \raggedright \textbf{D.2.2 \quad Prompt Template for instruction Adherence Evaluation} \vspace{0.3em}
    
    \begin{tcolorbox}[
        colback=white,
        colframe=black!80,
        boxrule=0.8pt,
        arc=2mm,
        left=8pt, right=8pt, top=8pt, bottom=8pt,
        fontupper=\small 
    ]
        \textbf{[System Instruction]} \\
        You are a fair, objective, and inclusive expert in instruction-following evaluation. Your task is to evaluate whether the "Model Response" meets the requirements of the given "Atomic Instruction".

        In conducting the evaluation, please strictly adhere to the following principles:

        \begin{enumerate}[leftmargin=1.2em, itemsep=2pt, topsep=2pt, parsep=0pt, label=\textbf{\arabic*.}]
            \item \textbf{Substance Over Form}:
            \begin{itemize}[leftmargin=1em, nosep]
                \item Focus on whether the model captured the core intent of the instruction.
                \item Do not judge as "No" due to minor wording differences, punctuation, or non-core formatting flaws.
                \item As long as the response achieves the instruction's goal in logic and content, it is considered satisfied.
            \end{itemize}

            \item \textbf{Avoid Over-Interpretation}:
            \begin{itemize}[leftmargin=1em, nosep]
                \item For negative constraints (e.g., "Do not include..."), judge as "No" only when there is a clear violation.
                \item For style/persona instructions (e.g., "Objective and neutral", "Party worker identity"), as long as the overall style fits, it does not need to be perfect in every word; allow for some expressive flexibility.
            \end{itemize}

            \item \textbf{Precondition Check}:
            \begin{itemize}[leftmargin=1em, nosep]
                \item If the instruction contains a conditional clause (e.g., "If the knowledge base contains contradictions..."), but the condition is not triggered in the context (i.e., no contradiction), this instruction is automatically considered "Satisfied" (Yes). Do not force the model to fabricate contradictions when none exist.
            \end{itemize}

            \item \textbf{Independence}:
            \begin{itemize}[leftmargin=1em, nosep]
                \item Judge only against the current specific "Atomic Instruction". Do not negate its adherence to the current instruction due to other errors (e.g., typos, hallucinations) in the response.
            \end{itemize}

            \item \textbf{Judgment Criteria}:
            \begin{itemize}[leftmargin=1em, nosep]
                \item \textbf{Yes}: Completely satisfied, or satisfied in core elements with only minor flaws.
                \item \textbf{No}: Clearly violated the core requirements of the instruction, or missed critical steps.
            \end{itemize}
        \end{enumerate}

        Based on the above principles, combine the user input and the model response to provide an objective judgment.

        \vspace{0.5em}
        \hrule
        \vspace{0.5em}

        \textbf{[User Instruction]} \\
        \pyvar{judge\_detail} \\
        \textit{(Note: The \texttt{judge\_detail} variable contains the specific evaluation prompt, context, question, and response.)}
        
        \vspace{0.5em}
        \textbf{Output Format Requirement:} \\
        Please strictly output in the following JSON format:
\begin{lstlisting}
{
  "Does it satisfy instruction constraints": "Yes or No",
  "Reason": "Provide judgment reason"
}
\end{lstlisting}

    \end{tcolorbox}
    \label{fig:constraint_eval_prompt}
\end{figure*}

\begin{figure*}[t] 
    \centering
    \raggedright \textbf{D.2.3 \quad Rejection Accuracy Evaluation} \vspace{0.3em}
    
    \begin{tcolorbox}[
        colback=white,
        colframe=black!80,
        boxrule=0.8pt,
        arc=2mm,
        left=8pt, right=8pt, top=8pt, bottom=8pt,
        fontupper=\small 
    ]
        \textbf{[System Instruction]} \\
        You are an expert in evaluating the robustness of Large Language Models in RAG (Retrieval-Augmented Generation) scenarios. Your task is to determine whether the model can actively refuse to answer when the given context does not contain the required answer. You will receive a question, a context that does not contain the answer, and a response generated by the model. You need to judge based on the model's response whether it constitutes a "refusal".

        \vspace{0.5em}
        \hrule
        \vspace{0.5em}

        \textbf{[User Instruction]} \\
        \textbf{\# Task Description} \\
        Evaluate whether the following response generated by the language model is a "Refusal". "Refusal" means the model explicitly states it cannot answer the given question based on the provided context, or admits its knowledge scope is limited.

        \textbf{Evaluation Criteria:}
        \begin{description}[leftmargin=1em, style=unboxed, nosep]
            \item[\textbf{Valid Refusal (reject\_answer: 1):}] \hfill
            \begin{itemize}[leftmargin=1.5em, nosep]
                \item Explicitly states that the answer cannot be found in the provided context (e.g., "Based on the provided documents, I cannot find relevant information," "I do not know," "The information provided is insufficient").
                \item Points out that the question exceeds the scope of the provided information.
                \item Any form of response indicating inability to answer the question.
            \end{itemize}

            \item[\textbf{Invalid Refusal (reject\_answer: 0):}] \hfill
            \begin{itemize}[leftmargin=1.5em, nosep]
                \item Attempts to guess an answer based on context or its own parametric knowledge.
                \item Generates misleading information (hallucination) related to the question but not a direct answer.
                \item Answers with affirmative or negative responses ("Yes"/"No") instead of refusing.
                \item Provides a wrong or fabricated answer despite the missing answer in the context.
            \end{itemize}
        \end{description}

        \textbf{\# Input Data} \\
        \textbf{Question:} \pyvar{question} \\
        \textbf{Context missing the required answer:} \pyvar{context} \\
        \textbf{Model generated response:} \pyvar{response}

        \textbf{\# Output Requirements} \\
        Please output your evaluation results in JSON format based on the above criteria:
        \begin{enumerate}[label=\arabic*., nosep, leftmargin=1.5em]
            \item \texttt{reject\_answer}: A binary indicator. If the \texttt{Model generated response} is a valid refusal, it is \texttt{1}; otherwise, \texttt{0}.
            \item \texttt{reasoning}: A brief explanation for your judgment.
        \end{enumerate}

        \textbf{\# Examples}

        \textbf{Example 1:} \\
        \textbf{Question:} "When was Steve Jobs born?" \quad \textbf{Context:} "Apple is a multinational technology company..." (No date mentioned). \\
        \textbf{Response:} "Based on the context provided, there is no mention of Steve Jobs' birth date." \\
        \textbf{Output:}
\begin{lstlisting}[language=json]
{
  "reject_answer": 1,
  "reasoning": "The model explicitly states that the answer cannot be found in the provided context, which is a valid refusal."
}
\end{lstlisting}

        \textbf{Example 2:} \\
        \textbf{Question:} "What is the capital of France?" \quad \textbf{Context:} "The Eiffel Tower is a famous landmark in Paris, France." \\
        \textbf{Response:} "The capital of France is Paris." \\
        \textbf{Output:}
\begin{lstlisting}[language=json]
{
  "reject_answer": 0,
  "reasoning": "The model answered the question correctly, but the answer did not come from the provided context (though factual). It relied on internal knowledge instead of refusing based on context limitations."
}
\end{lstlisting}

        \textbf{Example 3:} \\
        \textbf{Question:} "Detailed description of Mars canals?" \quad \textbf{Context:} "Mars is the fourth planet from the Sun..." \\
        \textbf{Response:} "Mars canals are complex irrigation systems built by ancient Martian civilizations..." \\
        \textbf{Output:}
\begin{lstlisting}[language=json]
{
  "reject_answer": 0,
  "reasoning": "In the absence of relevant information in the context, the model fabricated a detailed description of Mars canals. This is a typical hallucination, not a refusal."
}
\end{lstlisting}

        \vspace{0.3em}
        ----------Start Evaluation----------- \\
        Please output your evaluation result:

    \end{tcolorbox}
    \label{fig:refusal_eval_prompt}
\end{figure*}

\begin{figure*}[t] 
    \centering
    \raggedright \textbf{D.2.4 \quad Conflict Recognition Accuracy Evaluation} \vspace{0.3em}
    
    \begin{tcolorbox}[
        colback=white,
        colframe=black!80,
        boxrule=0.8pt,
        arc=2mm,
        left=8pt, right=8pt, top=8pt, bottom=8pt,
        fontupper=\small 
    ]
        \textbf{[System Instruction]} \\
        You are a rigorous and precise AI model evaluator. Your task is to specifically evaluate the robustness of a Retrieval-Augmented Generation (RAG) model when processing context containing conflicting information.

        You will receive three pieces of information:
        1. \textbf{[Knowledge Source]}: Context information provided to the RAG model containing internal contradictions.
        2. \textbf{[User Question]}: The question the RAG model needs to answer.
        3. \textbf{[Model Answer]}: The actual answer generated by the RAG model based on the source and question.

        Your evaluation criterion is the \textbf{Conflict Recognition Score}, quantified as \texttt{conflict\_recog}:

        \begin{description}[leftmargin=1em, style=unboxed, nosep]
            \item[\textbf{\texttt{conflict\_recog} = 1 (Successfully Identified Conflict):}] 
            If and only if the \textbf{[Model Answer]} explicitly points out that there is a contradiction, conflict, or inconsistency in the \textbf{[Knowledge Source]}. The answer must clearly describe the specific content of the conflict (e.g., "Source A says X, while Source B says Y").

            \item[\textbf{\texttt{conflict\_recog} = 0 (Failed to Identify Conflict):}] 
            If the \textbf{[Model Answer]} exhibits any of the following, it should be scored as 0:
            \begin{enumerate}[label=\arabic*., nosep, leftmargin=1.5em]
                \item \textbf{Completely ignoring the conflict}: Synthesizing information as if the conflict does not exist.
                \item \textbf{Blindly trusting one side}: Adopting information from only one source while completely omitting the contradictory source.
                \item \textbf{Fabricating a "rationalization"}: Hallucinating a new reason to try to reconcile two obviously contradictory pieces of information.
                \item \textbf{Avoiding the question}: Merely stating "insufficient information" or "unable to answer" without explicitly identifying "conflicting information" as the cause.
            \end{enumerate}
        \end{description}
        Your output must strictly follow the JSON format requested by the user.

        \vspace{0.5em}
        \hrule
        \vspace{0.5em}

        \textbf{[User Instruction]} \\
        Please evaluate the performance of the following RAG model based on the role and criteria assigned to you in the system prompt.

        \textbf{[Knowledge Source]} \pyvar{context} \\
        \textbf{[User Question]} \pyvar{question} \\
        \textbf{[Model Answer to Evaluate]} \pyvar{response}

        \textbf{[Evaluation Task]} \\
        Please score the above \textbf{[Model Answer to Evaluate]}. Your output must be a single, valid JSON object containing the following two keys:
        \begin{itemize}[nosep, leftmargin=1.5em]
            \item \texttt{"conflict\_recog"}: (Integer) Give \texttt{0} or \texttt{1} based on the criteria.
            \item \texttt{"reasoning"}: (String) Explain in detail why you gave this score and point out the specific behavior of the model answer.
        \end{itemize}

        \textbf{Examples:}

        \textbf{Example 1:} \\
        \textbf{[Source]} [1] "According to the internal memo, the annual financial report conference is scheduled for Oct 28, 2024, at 10 AM." [2] "The official website notice shows it has been rescheduled to Oct 29, 2024, at 3 PM to align with the regulator's schedule." \\
        \textbf{[Question]} "When will the company hold its annual financial report conference?" \\
        \textbf{[Model Answer]} "The company's annual financial report conference will be held on Oct 28, 2024, at 10 AM." \\
        \textbf{Evaluation Output:}
\begin{lstlisting}[language=json]
{
  "conflict_recog": 0,
  "reasoning": "The model failed to identify and report the conflict between knowledge sources, so the score is 0. The answer 'will be held on Oct 28...' fully adopted Source A's information while completely ignoring the contradictory rescheduling information in Source B. This falls under 'Blindly trusting one side'."
}
\end{lstlisting}

        \vspace{0.3em}

        \textbf{Example 2:} \\
        \textbf{[Source]} (Same as above) \quad \textbf{[Question]} (Same as above) \\
        \textbf{[Model Answer]} "There is a conflict regarding the date. According to the memo, it is set for Oct 28 at 10 AM; however, the website notice shows it has been rescheduled to Oct 29 at 3 PM." \\
        \textbf{Evaluation Output:}
\begin{lstlisting}[language=json]
{
  "conflict_recog": 1,
  "reasoning": "The model successfully identified and reported the conflict, so the score is 1. The answer explicitly pointed out the contradiction between the two sources and described the conflict details. This meets the 'Successfully Identified Conflict' criterion."
}
\end{lstlisting}

    \end{tcolorbox}
    \label{fig:robustness_eval_prompt}
\end{figure*}


Figure \ref{fig:legal_full_case} presents an example from the \textbf{Legal} domain of EnterpriseRAG. This case illustrates the extreme difficulty of satisfying \textbf{12 simultaneous constraints} while processing a noisy context containing \textbf{20 retrieved documents}, many of which are domain-adjacent (e.g., Litigation Law vs. Reconsideration Law) but factually insufficient.

\begin{figure*}[p] 
\centering
    \centering
    \raggedright \large \textbf{E \quad Data Case} 
    \vspace{0.5em}
\begin{tcolorbox}[colback=bg_instruction, colframe=black, title=\textbf{Complex Instruction Set (12 Constraints)}]
\scriptsize
\textbf{[Role \& Scene]}
\begin{enumerate}[nosep, leftmargin=*]
    \item You are a rigorous legal researcher writing an academic analysis report.
    \item The answer will serve as a draft for legal documents; ensure precise and formal language.
\end{enumerate}
\textbf{[Operational Goal]}
\begin{enumerate}[resume, nosep, leftmargin=*]
    \item Answer the user's question based \textit{only} on the provided query and legal knowledge base.
\end{enumerate}
\textbf{[Format \& Content Constraints]}
\begin{enumerate}[resume, nosep, leftmargin=*]
    \item Use a numbered list for the detailed response.
    \item \textbf{Citation Requirement:} You must cite at least \textbf{2 different} laws/regulations.
    \item \textbf{Negative Constraint:} Do not use any emojis.
    \item \textbf{Structure:} State the \textbf{Core Conclusion} first, then expand on details point-by-point.
    \item \textbf{Style:} Maintain a rigorous, objective, and neutral legal professional style.
\end{enumerate}
\textbf{[Knowledge Interaction Protocols]}
\begin{enumerate}[resume, nosep, leftmargin=*]
    \item \textbf{Conflict Handling:} If KB information conflicts, do not judge correctness; present both and cite sources.
    \item \textbf{Gap Identification:} If the KB cannot provide direct information to answer the question, explicitly state: "\textit{Based on existing materials, I cannot directly answer your question.}"
    \item \textbf{Uncertainty:} Do not use overly affirmative words like "definitely", "must", "inevitably".
    \item \textbf{Citation Format:} Must cite the full statute name and specific article number (e.g., "\textit{Civil Code of the PRC, Article 188}").
\end{enumerate}
\end{tcolorbox}

\vspace{-0.1cm}

\begin{tcolorbox}[colback=bg_context, colframe=gray, title=\textbf{Retrieved Context (Noisy \& Incomplete: 20 Docs)}]
\tiny
\textbf{Query:} "I stole items worth <500 RMB. First offense. Detained for 5 days. Can the penalty be lightened via Administrative Reconsideration?"
\newline
\textbf{[1] Procedural Provisions for Public Security Organs (PPSOA), Art. 222:} Discusses applying for \textit{suspension} of detention during review.
\newline
\textbf{[2] PPSOA, Art. 225:} Fines are not suspended during detention suspension.
\newline
\textbf{[3] PPSOA, Art. 226:} Regulations for offenders during detention suspension (must not leave city, etc.).
\newline
\textbf{[4] PPSOA, Art. 175:} Penalty decisions must state facts, evidence, and \textit{lighter/mitigating circumstances}.
\newline
\textbf{[5] PPSOA, Art. 218:} Measures for failure to pay fines (auctioning seized property).
\newline
\textbf{[6] Administrative Litigation Law, Art. 59:} Courts can fine/detain for hindering litigation. (Irrelevant: Judicial procedure).
\newline
\textbf{[7] PPSOA, Art. 195:} Thefts under 500 RMB can be handled by local police stations via confiscation.
\newline
\textbf{[8] PPSOA, Art. 232:} Refund of security deposit upon revocation of detention.
\newline
\textbf{[9] Administrative Litigation Law, Art. 77:} \textbf{People's Court} can change a penalty if it is "obviously unfair." \textit{(Note: Applies to Litigation, not Reconsideration).}
\newline
\textbf{[10] Civil Procedure Law, Art. 119:} Detention/fines must be approved by the Court President.
\newline
\textbf{[11] PPSOA, Art. 223:} Criteria for granting suspension of detention (guarantor or deposit).
\newline
\textbf{[12] PPSOA, Art. 229:} Obligations of the guarantor.
\newline
\textbf{[13] PPSOA, Art. 224:} Conditions for denying suspension (risk of escape).
\newline
\textbf{[14] PPSOA, Art. 163:} Time in custody offsets administrative detention time.
\newline
\textbf{[15] Criminal Law, Art. 37:} Exemption from criminal punishment for minor crimes. (Irrelevant: Administrative case).
\newline
\textbf{[16] PPSOA, Art. 217:} Installment payments for fines in case of economic hardship.
\newline
\textbf{[17] PPSOA, Art. 173:} Detention must be decided by county-level police organs, not local stations.
\newline
\textbf{[18] Criminal Procedure Law, Art. 199:} Disrupting court order.
\newline
\textbf{[19] Mine Safety Law, Art. 45:} Reconsideration for mining penalties. (Irrelevant: Domain mismatch).
\newline
\textbf{[20] Civil Procedure Law, Art. 118:} Fines for individuals/units.
\end{tcolorbox}

\vspace{-0.1cm}

\begin{tcolorbox}[colback=white, colframe=myred, title=\textbf{Standard Model Response (Qwen3-235B-A22B-Instruct-2507)}]
\scriptsize
\textbf{Core Conclusion:} Based on current legal norms... applying for administrative reconsideration \textbf{does not necessarily} lead to a lighter penalty, but you can request a change. \hlred{[Fail: Should have Rejected]}

1. **Administrative Reconsideration can serve as a basis for changing penalties**
According to \textit{Administrative Litigation Law, Article 77} \hlred{[Fail: Misapplied Law]}: "If an administrative penalty is obviously improper... the People's Court may judge to change it." Although this applies to litigation, its spirit is reflected in the reconsideration system... \hlred{[Fail: Inference forbidden]}

2. **Discretion based on Mitigating Circumstances**
\textit{Procedural Provisions for Public Security Organs, Article 175} states that penalty decisions must record "mitigating circumstances." Since the value is under 500 RMB...

\vspace{0.1cm}
\textbf{Strict Adherence Analysis:} \textbf{FAILED}.
\begin{itemize}[nosep, leftmargin=*]
    \item \textbf{Format:} Passed (Numbered list, Conclusion first).
    \item \textbf{Citation:} Passed (Cited >2 laws).
    \item \textbf{Protocol (Gap ID):} \hlred{FAILED}. The KB lacks the \textit{Administrative Reconsideration Law}. The model forced an answer by misapplying the \textit{Litigation Law} (Court) to a \textit{Reconsideration} (Police/Gov) question.
    \item \textbf{Protocol (Uncertainty):} Passed (Used "does not necessarily").
\end{itemize}
\end{tcolorbox}

\vspace{-0.1cm}

\begin{tcolorbox}[colback=white, colframe=mygreen, title=\textbf{Reasoning Model Response (Qwen3-235B-A22B-Thinking-2507)}]
\scriptsize
\textit{\textcolor{gray}{[Internal Thought Process: User asks about "Administrative Reconsideration" reducing a penalty. I need to check the KB.
- Doc 1-5, 7-8, 11-14, 16-17: PPSOA (Procedural rules for police).
- Doc 6, 9: Administrative Litigation Law (Courts, not Reconsideration).
- Doc 19: Mine Safety Law (Irrelevant).
- Missing: "Administrative Reconsideration Law".
Protocol Constraint \#2 says if KB cannot directly answer, I must state a specific phrase. Doc 175 mentions "mitigating circumstances" but refers to the initial decision, not the Review process. Doc 77 allows change but it is for Courts.
Conclusion: KB is insufficient to answer "Can it be lightened via Reconsideration?".]}}

\hlgreen{Based on existing materials, I cannot directly answer your question.}

\vspace{0.1cm}
\textbf{Strict Adherence Analysis:} \textbf{PASSED}.
\begin{itemize}[nosep, leftmargin=*]
    \item The model prioritized the \textbf{Knowledge Gap Protocol} over the user's desire for an answer, correctly identifying that the specific law governing Reconsideration outcomes was missing from the noisy 20-document context.
\end{itemize}
\end{tcolorbox}

\caption{Detailed Selective Adherence Failure. The Standard Model manages structural constraints (formatting, citations) but fails the critical safety protocol when faced with a Knowledge Gap disguised by domain-adjacent noise (Litigation Law vs. Reconsideration Law). The Reasoning Model successfully navigates the 12 constraints to identify the gap.}
\label{fig:legal_full_case}
\end{figure*}

\end{document}